\documentclass[11pt]{article}

\usepackage[preprint]{acl}

\usepackage{times}
\usepackage{latexsym}

\usepackage[T1]{fontenc}
\usepackage[utf8]{inputenc}

\usepackage{microtype}

\usepackage{inconsolata}

\usepackage{graphicx}

\usepackage{amsmath}
\usepackage{booktabs}
\usepackage{amssymb}
\usepackage{amsthm}
\usepackage[most]{tcolorbox}
\usepackage{makecell}

\theoremstyle{definition}

\theoremstyle{remark}

\usepackage[table]{xcolor}

\usepackage{enumitem}
\setlist[enumerate]{itemsep=2pt, parsep=0pt, topsep=4pt, leftmargin=*}

\usepackage{titlesec}
\titlespacing*{\paragraph}{0pt}{4pt plus 1pt minus 1pt}{0.5em}

\usepackage{pifont}
\newcommand{\cmark}{\ding{51}}%
\newcommand{\xmark}{\ding{55}}%

\usepackage{soul}
\usepackage{xcolor}
\definecolor{hlred}{RGB}{254,226,226}
\definecolor{hlgreen}{RGB}{220,252,231}
\sethlcolor{hlred}
\newcommand{\hlg}[1]{{\sethlcolor{hlgreen}\hl{#1}}}
\newcommand{\hlr}[1]{{\sethlcolor{hlred}\hl{#1}}}
\AtBeginDocument{
  \setlength{\textfloatsep}{6pt plus 2pt minus 2pt}
  \setlength{\floatsep}{6pt plus 2pt minus 2pt}
  \setlength{\intextsep}{6pt plus 2pt minus 2pt}
  \setlength{\dbltextfloatsep}{6pt plus 2pt minus 2pt}
  \setlength{\dblfloatsep}{6pt plus 2pt minus 2pt}
  \setlength{\abovecaptionskip}{3pt}
  \setlength{\belowcaptionskip}{0pt}
  \setlength{\abovedisplayskip}{5pt plus 2pt minus 2pt}
  \setlength{\belowdisplayskip}{5pt plus 2pt minus 2pt}
  \setlength{\abovedisplayshortskip}{1pt plus 1pt}
  \setlength{\belowdisplayshortskip}{3pt plus 1pt}
}

\title{Mitigating Rubric Interference in LLM 
Judges \\via On-Policy Self-Distillation}

\author{  Dingyao Yu$^{1,2}$ \quad
  Tong Zhang$^{2}$ \quad
  YuTao Mou$^{1,2}$ \quad
  Yunxiao Zhang$^{1,2}$ \\
  \textbf{Wei Ye}$^{1}$ \quad
  \textbf{Shikun Zhang}$^{1}$ \\
  \\
  $^{1}$ Peking University \\
  $^{2}$ Weixin AI, Tencent Inc \\
  \texttt{\{yudingyao,wye,shangsk\}@pku.edu.cn}}

\begin{document}
\maketitle
\begin{abstract}
LLM judges increasingly evaluate responses against fine-grained rubric checklists. When a sample requires multiple rubrics, current methods typically assess each in a separate inference call. Evaluating all rubrics in a single pass is a natural alternative with greater efficiency, but we find that it introduces \textit{rubric interference}: the verdict on one rubric shifts depending on which other rubrics are co-present. In a preliminary study, only one-third of samples receive fully consistent verdicts when evaluated under rubric sets of varying composition.
We develop a measurement framework that probes interference through four controlled operations: rubric set expansion, subsetting, reordering, and noise injection.
To mitigate interference without external supervision, we propose \textbf{Self-Anchored Rubric Alignment (SARA)}. SARA uses a model's own single-rubric judgments as stable anchors and aligns multi-rubric reasoning with these anchors through on-policy self-distillation.
We validate SARA on three datasets (HealthBench, FLASK, ResearchQA) and two model families (Qwen3, Llama-3.1). SARA consistently improves evaluation consistency while maintaining agreement with both base models and GPT-4.1 as a reference judge. Furthermore, the learned consistency transfers across datasets, confirming that SARA teaches a general capability rather than fitting dataset-specific patterns.
Our code is available at \href{https://anonymous.4open.science/r/SARA-22E2/}{https://anonymous.4open.science/r/SARA-22E2/}
\end{abstract}

\section{Introduction}

Large language models now serve as automated evaluators for
open-ended text generation, replacing or supplementing human
annotation in both benchmarking and reward modeling
\cite{NEURIPS2023_91f18a12, kim-etal-2024-prometheus}. Early
approaches rely on holistic scoring or pairwise comparison.
Holistic scoring collapses multidimensional quality into a
single number, sacrificing diagnostic granularity
\cite{chiang-lee-2023-large}. Pairwise comparison avoids
calibration issues but still provides no per-dimension feedback. These limitations have driven the
field toward rubric-based evaluation, which decomposes quality into specific, verifiable items for individual assessment
\cite{lee2025checkevalreliablellmasajudgeframework,
kim-etal-2024-prometheus, Pathak_2025}.

In tasks like HealthBench
\cite{arora2025healthbenchevaluatinglargelanguage} and ResearchQA
\cite{yifei2025researchqaevaluatingscholarlyquestion}, a single
sample requires evaluation against multiple rubrics. Two
strategies handle such multi-rubric sets.
\textit{Isolation evaluation} processes each rubric in a
separate inference call, producing independent verdicts.
\textit{Joint evaluation} assesses all rubrics in a single
pass, generating a unified structured report. Current practice
defaults to isolation
\cite{lee2025checkevalreliablellmasajudgeframework}.
Joint evaluation is attractive for its efficiency and ability
to produce coherent cross-rubric narratives, both desirable
when the judge serves as an online reward
signal in RLHF \cite{gunjal2026rubrics}. However, its
reliability remains largely unexamined.

\begin{figure}[t]
    \centering
    \includegraphics[width=\columnwidth]{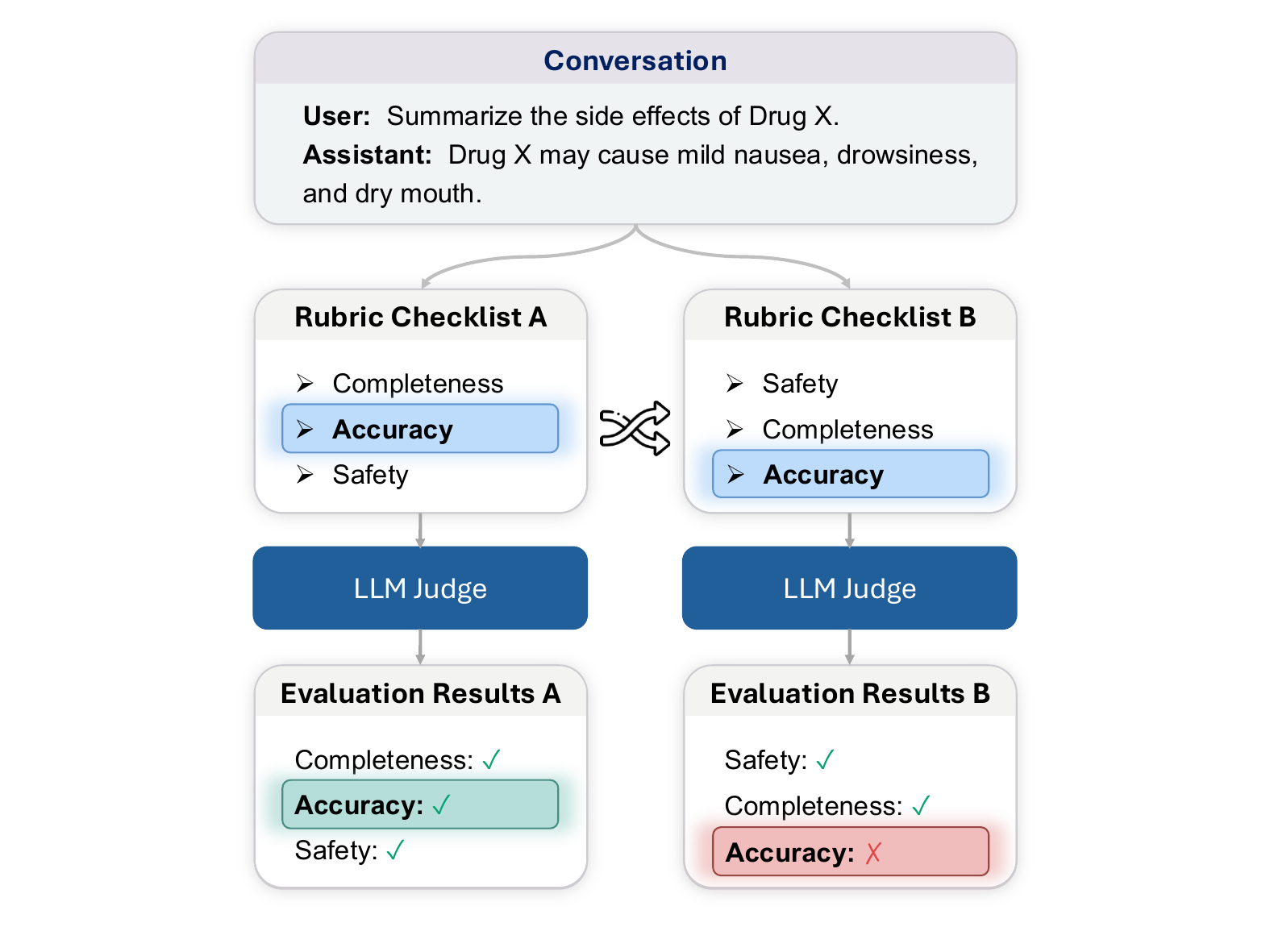}
    \caption{Rubric interference in joint evaluation. The same
    rubric receives different verdicts depending on the
    composition and ordering of co-present rubrics.}
    \label{fig:interference}
\end{figure}
  
For joint evaluation to reliably replace isolation calls, the
verdict on each rubric must depend only on the conversation and
that rubric. We find that this invariance does not hold. Joint
evaluation introduces \textit{rubric interference}: the verdict
on a rubric shifts depending on how the rubric set is composed.
Figure~\ref{fig:interference} illustrates one manifestation:
simply reordering the rubric list flips a verdict. Interference
also emerges when we expand from isolation to joint mode:
comparing each rubric's isolation verdict against its joint
verdict on HealthBench, even Qwen3-32B achieves only 36\%
sample-level exact match. This reflects systematic interference
rather than generation noise: all models achieve $\geq$0.98
self-agreement across repeated evaluations with different
random seeds.

Rubric interference is systematic and reproducible. It differs
from known judge biases such as position, verbosity, and
self-enhancement
\cite{wang-etal-2024-large-language-models-fair,
wu-aji-2025-style, panickssery2024llmevaluatorsrecognizefavor},
which concern how a judge ranks different \textit{responses}.
Rubric interference concerns how the judge handles different
\textit{evaluation contexts} for the same response.

A key observation enables our solution. While joint evaluation
is unreliable, the same model evaluating each rubric \textit{in
isolation} produces stable judgments that require no external
annotation. The gap between isolation and joint verdicts
reveals where interference distorts reasoning, providing a
training signal without external supervision.

We propose \textbf{Self-Anchored Rubric Alignment (SARA)},
which aligns joint-mode reasoning with the model's own
isolation judgments through on-policy self-distillation. SARA
generates a multi-rubric evaluation, parses it into per-rubric
segments, and distills from a slowly updated teacher that
evaluates each segment in isolation context. A divergence loss
corrects analysis segments where interference distorts
reasoning; a preservation loss maintains verdict structure.

Our contributions are as follows:
\begin{enumerate}
\item We identify rubric interference as a systematic failure mode in multi-rubric LLM evaluation and develop a measurement framework with four operations to quantify it across binary and scored formats.
\item We propose SARA, a self-distillation method that mitigates rubric interference by anchoring joint-mode reasoning to the model's own isolation judgments, requiring no external supervision.
\item We validate SARA on three datasets and two model families. SARA improves consistency while preserving evaluation quality relative to both base models and GPT-4.1.
\end{enumerate}

\section{Related Work}

\paragraph{LLM-as-a-Judge.}
LLM judges have become standard evaluation infrastructure
\cite{NEURIPS2023_91f18a12,
li2024llmsasjudgescomprehensivesurveyllmbased}. Early work
showed that GPT-4 matches inter-annotator agreement on
MT-Bench, justifying LLM judges at scale
\cite{NEURIPS2023_91f18a12}. Chain-of-thought prompting further
improves correlation with human ratings \cite{liu2023geval},
and task-specific checklists from real users have moved
evaluation toward structured per-sample criteria
\cite{lin2024wildbench}. More recently, rubric-based evaluation
decomposes quality into fine-grained, verifiable items and
shows improved reproducibility across judge models
\cite{lee2025checkevalreliablellmasajudgeframework,
kim-etal-2024-prometheus, Pathak_2025}. Benchmarks such as
HealthBench \cite{arora2025healthbenchevaluatinglargelanguage}
and ResearchQA
\cite{yifei2025researchqaevaluatingscholarlyquestion} adopt
per-sample rubric sets as part of the evaluation protocol. Our
work identifies rubric interference as a reliability concern
specific to joint multi-rubric assessment.

\paragraph{Biases and reliability of LLM judges.}
LLM judges exhibit systematic biases including position bias,
verbosity bias, and self-enhancement bias
\cite{wang-etal-2024-large-language-models-fair,
wu-aji-2025-style, panickssery2024llmevaluatorsrecognizefavor}.
Position bias in particular is stable and model-specific rather
than a sampling artifact, persisting across 12 judge models and
over 100K evaluation instances \cite{NEURIPS2023_91f18a12}.
Sensitivity to prompt complexity and a general leniency
tendency further compromise reliability
\cite{thakur2024judging}. Even GPT-4o performs near chance on
challenging response pairs \cite{tan2024judgebench}.
Calibration strategies have been proposed to address these
issues \cite{li-etal-2025-calibraeval,
li2026evaluatingscoringbiasllmasajudge}. All these findings
concern \textit{inter-response} biases---how the judge
compares different responses. Rubric interference is
orthogonal: the judgment on a single rubric shifts depending
on the co-evaluated set, with the response fixed.

\paragraph{Prompt sensitivity and evaluation consistency.}
Beyond judge-specific biases, LLMs exhibit broad prompt
sensitivity: formatting changes alone cause up to 76-point
accuracy swings \cite{sclar2024quantifying}, motivating
multi-prompt evaluation as a methodological default
\cite{mizrahi2024state}. Paraphrase-based consistency
frameworks measure whether equivalently phrased queries elicit
the same factual answer \cite{elazar2021measuring}. These works
study sensitivity of LLMs as \textit{task-solvers}. We study
sensitivity of LLMs as \textit{judges}: how context
changes the label assigned to another model's output.

\paragraph{On-policy distillation.}
On-policy distillation trains a student on self-sampled
trajectories with per-token teacher guidance, mitigating
distribution shift
\cite{agarwal2024onpolicydistillationlanguagemodels,
gu2024minillm}. Self-distillation removes the need for a
separate teacher: distilling a model into itself can improve
generalization \cite{pmlr-v80-furlanello18a}. In the LLM
setting, models have been trained to distinguish their own
outputs from references
\cite{chen2024selfplayfinetuningconvertsweak}, to serve as
their own reward function \cite{pmlr-v235-yuan24d}, and to
teach themselves by conditioning on privileged information
\cite{lu2025opsd}. SARA shares this on-policy
self-distillation principle but targets a different goal:
aligning joint-mode with isolation-mode evaluation. The
privileged information is not a stronger model but the absence
of interfering rubrics.

\section{Problem Definition}

\subsection{Joint Rubric Evaluation}

We consider a judge model $\mathcal{M}$ that evaluates a conversation $c$ against a set of rubrics $R = \{r_1, \ldots, r_n\}$. For each rubric, the judge assigns a verdict $v_i$ from a discrete value set $\mathcal{V}$. We study binary ($\mathcal{V} = \{0, 1\}$) and scored (e.g. $\mathcal{V} = \{1, \ldots, 5\}$) evaluation; the framework extends to any ordinal scale.

In \textbf{isolation evaluation}, the judge processes each rubric in a separate inference call, producing verdict $v_i^{\text{iso}}$ from $\mathcal{M}(c, \{r_i\})$. In \textbf{joint evaluation}, the judge assesses all rubrics in a single pass, producing verdict $v_i^{\text{jnt}}$ from $\mathcal{M}(c, R)[r_i]$. Joint evaluation is more efficient, but the verdict $v_i^{\text{jnt}}$ may depend on the other rubrics in $R$, introducing rubric interference.

An ideal judge produces the same verdict for rubric $r$ regardless of which other rubrics accompany it: 

\vspace{-6pt}
\begin{equation}
  \mathcal{M}(c, R)[r] = \mathcal{M}(c, R')[r]
  \quad \forall\; R, R' \ni r
\end{equation}

The rest of this section describes how we test whether real judges satisfy it.

\subsection{Testing for Rubric Interference}

We test consistency by changing the rubric set in controlled ways and checking whether verdicts on shared rubrics remain stable. Table~\ref{tab:perturbation} summarizes four operations, each targeting a different interference mechanism.

\begin{table}[t]
\centering
\small
\setlength{\tabcolsep}{4pt}
\begin{tabular}{lll}
\toprule
\textbf{Operation} & \textbf{What changes} & \textbf{What we compare} \\
\midrule
Expansion
  & Set size: $1 \to n$
  & $v_i^{\text{iso}}$ vs $v_i^{\text{jnt}}$ \\
Subsetting
  & Set size: $m \to n$
  & $v_i$ in small vs large set \\
Reordering
  & Rubric order
  & $v_i$ across permutations \\
Noise
  & Add unrelated rubrics
  & $v_i$ before vs after noise \\
\bottomrule
\end{tabular}
\caption{Four operations for testing rubric interference.
Each changes the rubric set while keeping the conversation
and target rubric fixed.}
\label{tab:perturbation}
\end{table}

\textbf{Expansion} is our primary test: we compare each
rubric's isolation verdict against its joint verdict. We report
\textit{overall consistency} (full set) and
\textit{consistency-at-$K$} (random subsets of size $K$) to
trace how interference grows with rubric count.
\textbf{Subsetting} generalizes expansion by comparing two
nested subsets of arbitrary size rather than anchoring on
isolation; this tests whether interference accumulates
monotonically but requires sampling valid superset. \textbf{Reordering} evaluates
the same set under multiple permutations, isolating positional
effects from content-based interference. \textbf{Noise
injection} appends rubrics from unrelated conversations,
testing whether the model can ignore irrelevant context.

For all operations, we report rubric-level agreement and
Cohen's $\kappa$, plus sample-level exact match (EM). Full
metric definitions are in Appendix~\ref{app:metrics}.

%% ================================================================
\section{Self-Anchored Rubric Alignment}

\begin{figure*}[t!]
\centering
\includegraphics[width=0.95\textwidth]{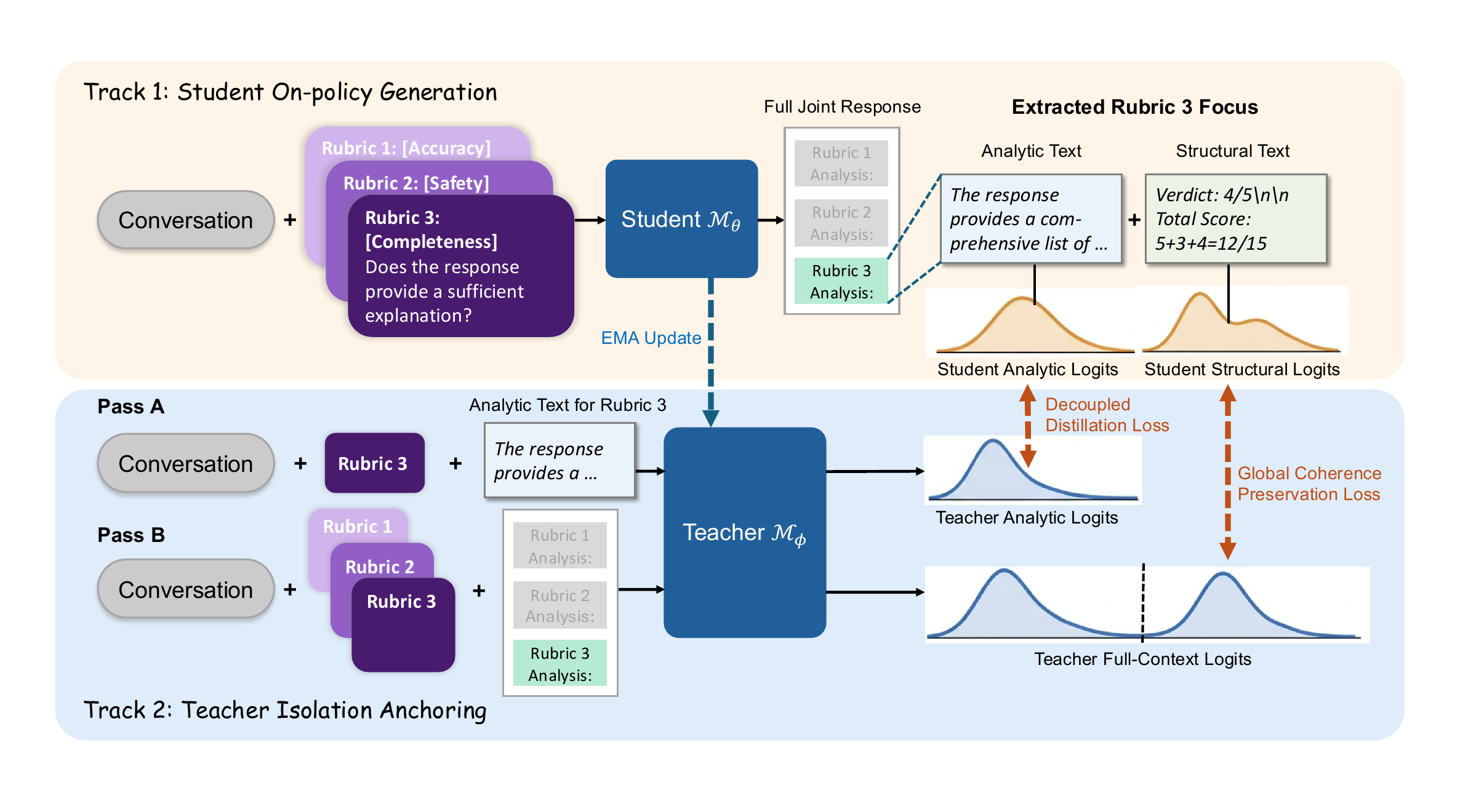}
\caption{The SARA training pipeline. The student generates a
joint evaluation (top), which is parsed into per-rubric
segments. The teacher evaluates each segment under isolation
context (bottom), producing interference-free anchor logits.
The distillation loss aligns student and teacher logits on
analysis segments; the preservation loss maintains structural
tokens.}
\label{fig:sara}
\end{figure*}

\subsection{Overview}

SARA trains a judge to resist rubric interference through on-policy self-distillation. The core idea is simple: the same model that struggles with joint evaluation judges each rubric reliably in isolation. These isolation judgments are stable and require no external annotation, making them natural training anchors. 

At each training step, the student model generates a joint evaluation and parses it into per-rubric segments. A slowly updated copy of the student (the teacher) then evaluates each segment in isolation context, producing interference-free logits. A divergence loss aligns the student's joint-mode logits with these anchors on analysis segments, while a preservation loss maintains the structural tokens (verdicts, formatting). Figure~\ref{fig:sara} shows the pipeline.

\subsection{On-Policy Generation and Segment Extraction}

At each training step, the student $\mathcal{M}_\theta$
generates a multi-rubric evaluation for conversation $c$ and rubric
set $R$:
\begin{equation}
  y \sim \mathcal{M}_\theta(\cdot \mid c, R)
    = [a_1, s_1, \;\ldots,\; a_n, s_n]
\end{equation}
Here $a_i$ is the analysis segment for rubric $r_i$ and $s_i$ is the structural segment (verdict, formatting, transition tokens) . Because $y$ is sampled from the current model at each step
(on-policy), the training distribution tracks the model's
evolving behavior. 

We parse $y$ into per-rubric segments using the structured
output format (Appendix~\ref{app:prompt}). This produces an
\textit{analysis mask}
$\mathbf{m}^{\text{ana}} \in \{0,1\}^{|y|}$ that marks tokens
in all $a_i$ segments, and a complementary \textit{structure
mask} $\mathbf{m}^{\text{str}} = \mathbf{1} -
\mathbf{m}^{\text{ana}}$ that covers all $s_i$ segments.

Segment extraction assumes instruct-mode outputs with a predictable rubric-analysis-verdict structure. Thinking-mode models require model-specific parsing, discussed in Appendix~\ref{app:thinking}. 

\subsection{Isolation-Anchored Distillation}

\paragraph{Teacher model.}
The teacher $\mathcal{M}_\phi$ is a slowly updated copy of the
student. After each training step:
\begin{equation}
  \phi \leftarrow \beta_{\text{ema}}\,\phi
    + (1 - \beta_{\text{ema}})\,\theta
\end{equation}
The high decay rate ($\beta_{\text{ema}}$ close to 1) provides
a stable reference throughout training. Using the teacher
rather than the student itself as anchor avoids a degenerate
loop where drifting judgments reinforce interference.

Two considerations motivate this design over using the
student's own isolation logits directly. First, EMA smooths
the reference temporally and prevents the training signal
from oscillating at each update. Second, it breaks a
potential degenerate loop: if the student anchored on its own
isolation pass, any systematic bias would reinforce itself
without correction.

\paragraph{Distillation loss.}
For each analysis segment $a_i$, we compare two forward passes
over the same tokens: the student $\mathcal{M}_\theta$ under
the full rubric set $R$, and the teacher $\mathcal{M}_\phi$
under isolation context $\{r_i\}$:
\begin{equation}
\begin{aligned}
  \mathbf{p} &= \mathcal{M}_\theta\!\left(
    y \;\middle|\; c,\; R\right) \\
  \hat{\mathbf{q}}^{(i)} &= \mathcal{M}_\phi\!\left(
    a_i \;\middle|\; c,\;\{r_i\}\right)
\end{aligned}
\end{equation}
Here $\mathbf{p}$ denotes the student's logits over its
complete joint-mode output $y$, conditioned on conversation $c$
and all rubrics $R$. The teacher logits
$\hat{\mathbf{q}}^{(i)}$ cover only the $i$-th analysis
segment $a_i$, conditioned on the single rubric $r_i$. The
student sees all rubrics and may exhibit interference; the
teacher sees only the target rubric and produces
interference-free logits. We align them across all analysis
tokens:
\begin{equation}
  \mathcal{L}_{\text{distill}} = \frac{1}{|\mathcal{A}|}
    \sum_{i=1}^{n}\sum_{t \in \mathcal{A}_i}
    D_{\text{JSD}}^{(\beta_d)}\!\left(
      \mathbf{p}_t \;\|\; \hat{\mathbf{q}}^{(i)}_t\right)
\end{equation}
where $\mathcal{A}_i$ indexes token positions in segment $a_i$,
$\mathcal{A} = \bigcup_i \mathcal{A}_i$ is the full analysis
token set marked by $\mathbf{m}^{\text{ana}}$.

\paragraph{Preservation loss.}
Structural tokens (verdicts, formatting, transitions) need a
separate signal. We align the student's logits with the
teacher's full-context logits, both conditioned on the
complete rubric set $R$:
\begin{equation}
  \mathbf{q} = \mathcal{M}_\phi\!\left(
    y \;\middle|\; c,\; R\right)
\end{equation}
Here $\mathbf{q}$ denotes the teacher's logits over the same
output $y$ under the same full context. The only difference
from $\mathbf{p}$ is the model weights ($\phi$ vs $\theta$).
We compute the preservation loss over structural token
positions $\mathcal{S}$ marked by $\mathbf{m}^{\text{str}}$:
\begin{equation}
  \mathcal{L}_{\text{preserve}} = \frac{1}{|\mathcal{S}|}
    \sum_{t \in \mathcal{S}}
    D_{\text{JSD}}^{(\beta_p)}\!\left(
      \mathbf{p}_t \;\|\; \mathbf{q}_t\right)
\end{equation}
We use full-context teacher logits here because structural
tokens coordinate across all rubrics and have no meaningful
isolation equivalent.

\paragraph{Total loss.}
\begin{equation}
  \mathcal{L} = \mathcal{L}_{\text{distill}}
    + \alpha \, \mathcal{L}_{\text{preserve}}
\end{equation}

Since structural tokens are far fewer than analysis tokens,
each structural token already receives higher per-token weight
at equal $\alpha$. We set $\alpha{=}0.4$ across all
experiments.

\subsection{Rubric Context Augmentation}

To expose the model to diverse interference patterns, the data collator applies two augmentations to each rubric set $R$ at every training step:

\begin{itemize}[itemsep=2pt, parsep=0pt, topsep=4pt]
\item \textbf{Rubric shuffling.} We randomly permute the rubric
order at each step, preventing the model from memorizing
position-specific patterns.

\item \textbf{Noise injection.} With probability
$p_{\text{noise}}$, we append
$k$ rubrics from unrelated
conversations, training the model to ignore irrelevant
criteria.
\end{itemize}

These two augmentations target the interference sources in
Table ~\ref{tab:perturbation}. Shuffling addresses positional
sensitivity and noise injection addresses content-based
distraction.

\section{Experiments}
\label{sec:exp}

\subsection{Setup}

We evaluate on three datasets spanning distinct domains and
scoring formats (Table~\ref{tab:datasets}). HealthBench
(HB) contains medical conversations evaluated against
per-sample rubric checklists with binary met/unmet verdicts.
Rubric counts range from 2 to 32 per sample, making it the
most demanding test for interference.
FLASK evaluates general-purpose instruction-following
across 3 fixed skill dimensions, each scored on a 1--5 scale.
ResearchQA (RQA) evaluates scientific question
answering against per-sample rubrics scored on a 0--4 scale.
We sampled 1000 instances from the full dataset.

\begin{table}[t]
\centering
\small
\setlength{\tabcolsep}{4pt}
\begin{tabular}{lccc}
\toprule
& \textbf{HB} & \textbf{FLASK} & \textbf{RQA} \\
\midrule
Domain         & Medical   & General    & Scientific \\
Scoring        & Binary    & 1--5       & 0--4 \\
Rubrics/sample & 2--32     & 3 (fixed)  & 1--8 \\
Avg rubrics    & 11.5      & 3.0        & 7.5 \\
Samples        & 500       & 1740       & 1000 \\
\bottomrule
\end{tabular}
\caption{Dataset statistics. HB\,=\,HealthBench,
RQA\,=\,ResearchQA.}
\label{tab:datasets}
\end{table}

\paragraph{Models and baselines.}
We apply SARA to Qwen3-8B, Qwen3-14B, Qwen3-32B, and
Llama-3.1-8B-Instruct. As a baseline, we compare against
\textbf{SFT\textsubscript{iso}}: standard supervised
fine-tuning on isolation outputs concatenated into joint
format. This represents the most direct approach to teaching joint-mode behavior from
isolation references. Each dataset is split 8:1:1 for
training, validation, and test.

\paragraph{Training details.}
SARA uses full-parameter fine-tuning with EMA decay 0.999,
symmetric JSD ($\beta_d{=}0.5$), KL preservation
($\beta_p{=}1.0$, $\alpha{=}0.4$), and vLLM greedy decoding
for on-policy generation. We select checkpoints by shuffle
consistency on the validation set.

\paragraph{Evaluation.}
All evaluations use greedy decoding with fixed
random seeds. We cap rubrics at 10 per sample. For shuffle
tests, we average over 5 random permutations. We report
rubric-level agreement (Agr), Cohen's $\kappa$ (weighted for
scored formats), and sample-level exact match (EM).

\paragraph{Infrastructure.}
All experiments run on 8$\times$ NVIDIA H20 GPUs with vLLM
for batched inference.

\begin{table*}[t]
\centering
\small
\begin{tabular}{l ccc ccc ccc}
\toprule
& \multicolumn{3}{c}{\textbf{HealthBench}}
& \multicolumn{3}{c}{\textbf{ResearchQA}}
& \multicolumn{3}{c}{\textbf{FLASK}} \\
\cmidrule(lr){2-4} \cmidrule(lr){5-7} \cmidrule(lr){8-10}
& Agr{\scriptsize$\uparrow$} & $\kappa${\scriptsize$\uparrow$} & EM{\scriptsize$\uparrow$}
& Agr{\scriptsize$\uparrow$} & $\kappa${\scriptsize$\uparrow$} & EM{\scriptsize$\uparrow$}
& Agr{\scriptsize$\uparrow$} & $\kappa${\scriptsize$\uparrow$} & EM{\scriptsize$\uparrow$} \\
\midrule
Qwen3-8B
  & .819 & .630 & .22
  & .774 & .665 & .22
  & .720 & .663 & .41 \\
\quad\textit{+SARA}
  & \textbf{.874} & \textbf{.736} & \textbf{.34}
  & \textbf{.914} & \textbf{.868} & \textbf{.59}
  & \textbf{.810} & \textbf{.801} & \textbf{.52} \\
\midrule
Qwen3-14B
  & .874 & .747 & .34
  & .768 & .658 & .23
  & .722 & .665 & .41 \\
\quad\textit{+SARA}
  & \textbf{.903} & \textbf{.806} & \textbf{.38}
  & \textbf{.900} & \textbf{.879} & \textbf{.52}
  & \textbf{.803} & \textbf{.802} & \textbf{.55} \\
\midrule
Qwen3-32B
  & .887 & .775 & .36
  & .720 & .651 & .25
  & .701 & .652 & .41 \\
\quad\textit{+SARA}
  & .881 & .762 & .36
  & \textbf{.880} & \textbf{.851} & \textbf{.42}
  & \textbf{.836} & \textbf{.822} & \textbf{.59} \\
\midrule
Llama-3.1-8B
  & .733 & .467 & .08
  & .711 & .608 & .12
  & .609 & .637 & .24 \\
\quad\textit{+SARA}
  & \textbf{.830} & \textbf{.652} & \textbf{.18}
  & \textbf{.853} & \textbf{.800} & \textbf{.34}
  & \textbf{.722} & \textbf{.760} & \textbf{.41} \\
\bottomrule
\end{tabular}
\caption{Overall consistency (isolation vs.\ joint mode).
Agr\,=\,rubric-level agreement, $\kappa$\,=\,Cohen's kappa
(weighted for scored formats), EM\,=\,sample-level exact match.}
\label{tab:main}
\end{table*}

\subsection{Main Results}
\label{sec:main_results}

Table~\ref{tab:main} presents overall consistency between
isolation and joint mode. SARA delivers consistent improvements
across model scales, architectures, and evaluation domains.

The largest gains appear on ResearchQA: Qwen3-8B's EM nearly triples (.22$\to$.59),
and its $\kappa$ jumps from .665 to .868. ResearchQA combines
moderate rubric counts with per-sample rubric
definitions, creating ample opportunity for cross-rubric
contamination. On HealthBench, binary verdicts set a higher
baseline and lower ceiling, yet improvements remain steady
across models up to 14B. On FLASK, only 3 fixed rubrics limit
the surface area for interference, yet gains persist (e.g.,
Qwen3-32B: EM .41$\to$.59), confirming that even minimal
rubric sets trigger measurable interference.

SARA benefits weaker models more (Llama-3.1-8B: +14.2\% Agr
on ResearchQA), consistent with our hypothesis that stronger
models have implicitly learned partial invariance through
scale. Yet even Qwen3-32B improves substantially on scored
formats, suggesting scale alone does not fully resolve
interference. EM gains are disproportionately large relative
to Agr across all settings (e.g., Qwen3-8B on ResearchQA:
Agr +.140 but EM +.370). This asymmetry reveals that baseline
interference is not concentrated on a few hard rubrics but
scattered across many---a single flipped rubric per sample
suffices to break exact match, and SARA systematically
corrects these distributed errors. On scored datasets, SARA
also narrows the gap between Agr and $\kappa$ (Qwen3-32B on
ResearchQA: .720/.651$\to$.880/.851), indicating that
remaining disagreements concentrate at adjacent scores
($\pm$1) rather than large deviations---interference is not
only less frequent but less severe.

\subsection{Consistency Without Quality Degradation}
\label{sec:quality}

\begin{table}[ht]
\centering
\small
\setlength{\tabcolsep}{3pt}
\begin{tabular}{l ccc cc}
\toprule
& \multicolumn{3}{c}{\textbf{Consistency}}
& \textbf{Iso Agr w/}
& \textbf{Jnt Agr w/} \\
& Agr{\scriptsize$\uparrow$} & $\kappa${\scriptsize$\uparrow$} & EM{\scriptsize$\uparrow$}
& Base$^*${\scriptsize$\uparrow$}
& Base$^*${\scriptsize$\uparrow$} \\
\midrule
Qwen3-14B
  & .768 & .658 & .23 & -- & -- \\
+SFT\textsubscript{iso}
  & \textbf{.909} & .872 & \textbf{.54} & .892 & .777 \\
+SARA
  & .900 & \textbf{.879} & .52 & \textbf{.925} & \textbf{.806} \\
\bottomrule
\end{tabular}
\caption{SFT\textsubscript{iso} vs.\ SARA on ResearchQA. Left: consistency between evaluation modes. Right: agreement with the untrained base model's outputs.}
\label{tab:sft_compare}
\end{table}

\paragraph{Why not naive SFT?}
A natural baseline is to directly fine-tune on isolation
outputs concatenated into joint format
(SFT\textsubscript{iso}: 5 random-seed isolation samples per
rubric as synthetic joint targets).
Table~\ref{tab:sft_compare} compares this approach against
SARA on Qwen3-14B (ResearchQA). Beyond consistency,
we measure behavioral drift: ``Iso.\ Agr with Base$^*$''
reports how much a trained model's isolation outputs agree with
the untrained base; ``Joint Agr with Base$^*$'' does the same for
joint outputs.

SFT\textsubscript{iso} achieves slightly higher iso--joint
consistency (Agr .909 vs .900, EM .54 vs .52), but drifts
further from the original model's behavior in both modes
(Iso: .892 vs .925; Joint: .777 vs .806). SFT forces the
model to \emph{reproduce} fixed target outputs, inadvertently
shifting its judgment distribution; SARA teaches the model to
\emph{reason independently per rubric}, preserving its
original evaluation behavior while reducing interference.

\begin{table}[ht]
\centering
\small
\setlength{\tabcolsep}{4pt}
\begin{tabular}{l cc cc}
\toprule
& \multicolumn{2}{c}{\textbf{Iso.\ Agr w/}}
& \multicolumn{2}{c}{\textbf{Joint Agr w/}} \\
\cmidrule(lr){2-3} \cmidrule(lr){4-5}
& Base$^*$ & GPT-4.1 & Base$^*$ & GPT-4.1 \\
\midrule
Qwen3-8B       & --   & .729 & --   & .773 \\
\quad\textit{+SARA} & .921 & .706 & .874 & .731 \\
\midrule
Qwen3-14B      & --   & .801 & --   & .795 \\
\quad\textit{+SARA} & .912 & .806 & .896 & .806 \\
\midrule
Qwen3-32B      & --   & .830 & --   & .815 \\
\quad\textit{+SARA} & .896 & .823 & .894 & .819 \\
\midrule
Llama-3.1-8B   & --   & .702 & --   & .726 \\
\quad\textit{+SARA} & .839 & .700 & .755 & .755 \\
\bottomrule
\end{tabular}
\caption{Evaluation quality on HealthBench. Base$^*$:
agreement with untrained base. GPT-4.1: external reference
judge.}
\label{tab:quality}
\end{table}

\paragraph{Does SARA preserve evaluation quality?}
Table~\ref{tab:quality} confirms across all models on
HealthBench that SARA's consistency gains do not come at the
expense of judgment quality. Agreement with the untrained base
exceeds .84 in all settings, confirming minimal behavioral
drift. GPT-4.1 agreement is preserved or improved (Qwen3-14B
joint: .795$\to$.806; Llama-3.1-8B joint: .726$\to$.755):
when isolation judgments are already reasonable, reducing
interference also improves absolute accuracy.

\subsection{Interference Under Controlled Perturbations}
\label{sec:con_k}

We apply the perturbation framework from
Table~\ref{tab:perturbation} to trace how each interference source
affects model behavior, and how SARA mitigates it.

\begin{figure}[ht]
\centering
\includegraphics[width=\columnwidth]{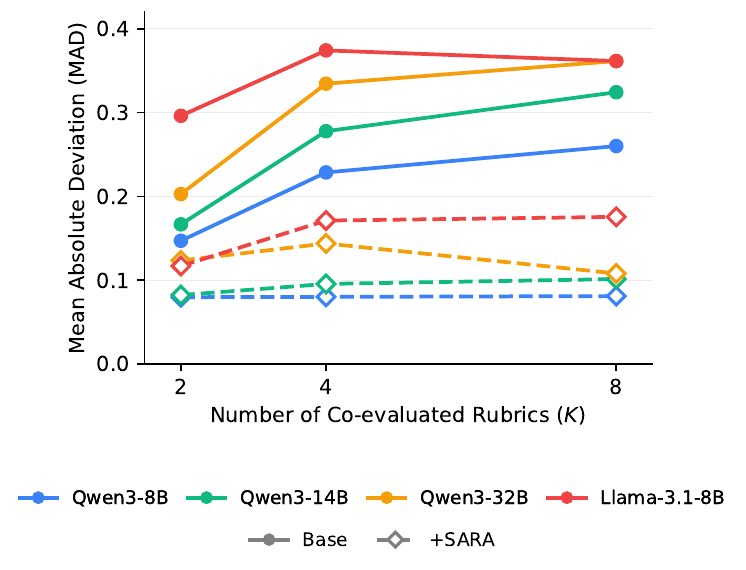}
\caption{Mean Absolute Deviation (MAD) between isolation and
joint verdicts on ResearchQA as co-evaluated rubric count $K$
grows.}
\label{fig:con_k}
\end{figure}

\paragraph{Scaling with rubric count.}
Figure~\ref{fig:con_k} traces MAD on ResearchQA as $K$ grows
from 2 to 8. For all base models, MAD rises steeply with $K$:
Qwen3-32B climbs from .203 at $K{=}2$ to .362 at $K{=}8$,
nearly doubling. SARA flattens these curves. Qwen3-8B's MAD
stays nearly constant (.080 at $K{=}2$, .080 at $K{=}4$, .081
at $K{=}8$), indicating that the model evaluates each rubric
independently of set size. Across all models, SARA compresses
the MAD range from .147--.374 (base) to .080--.176 (trained)
and eliminates the upward trend with $K$.

\begin{figure}[ht]
\centering
\includegraphics[width=\columnwidth]{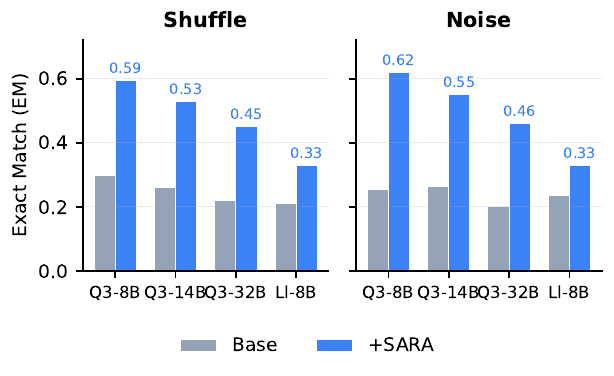}
\caption{Shuffle invariance and noise robustness on
ResearchQA (EM).}
\label{fig:shuffle_noise_sum}
\end{figure}

\paragraph{Shuffle and noise robustness.}
Figure~\ref{fig:shuffle_noise_sum} reports EM under random
rubric permutation (shuffle) and irrelevant rubric injection
(noise) on ResearchQA. Under shuffling, Qwen3-14B achieves
only .262 EM; under noise, Qwen3-32B drops to .200. SARA
roughly doubles shuffle EM across all models (e.g., Qwen3-8B:
.298$\to$.594) and yields similar noise gains (Qwen3-8B:
.255$\to$.620). Full results across all datasets are in
Appendix~\ref{app:shuffle_noise}.

\subsection{SARA Learns Transferable Consistency}
\label{sec:transfer}

To confirm that SARA teaches a general capability rather than fitting dataset-specific patterns, we evaluate Qwen3-14B trained on ResearchQA (SARA\textsubscript{rqa}) directly on HealthBench and FLASK without further fine-tuning (Table~\ref{tab:transfer}).

\begin{table}[ht]
\centering
\small
\setlength{\tabcolsep}{4pt}
\begin{tabular}{l ccc}
\toprule
\multicolumn{4}{c}{\textbf{HealthBench} (Agr / $\kappa$ / EM)} \\
\midrule
Qwen3-14B                     & .874 & .747 & .34 \\
\quad +SARA                    & \textbf{.903} & \textbf{.806} & \textbf{.38} \\
\quad +SARA\textsubscript{rqa} & .890 & .779 & .34 \\
\midrule
\multicolumn{4}{c}{\textbf{FLASK} (Agr / $\kappa$ / EM)} \\
\midrule
Qwen3-14B                     & .722 & .665 & .41 \\
\quad +SARA                    & .803 & \textbf{.802} & \textbf{.55} \\
\quad +SARA\textsubscript{rqa} & \textbf{.820} & .800 & \textbf{.55} \\
\bottomrule
\end{tabular}
\caption{Cross-dataset transfer. SARA\textsubscript{rqa}
(trained on ResearchQA) evaluated zero-shot on HealthBench
and FLASK. }
\label{tab:transfer}
\end{table}

SARA\textsubscript{rqa} improves over the base model on both
target datasets. On FLASK it even outperforms the in-domain
SARA model in Agr (.820 vs .803). This confirms that SARA
learns a domain-agnostic ability to decouple per-rubric
reasoning from contextual interference.

\subsection{Analysis}
\label{sec:analysis}

\begin{figure}[ht]
\centering
\includegraphics[width=\columnwidth]{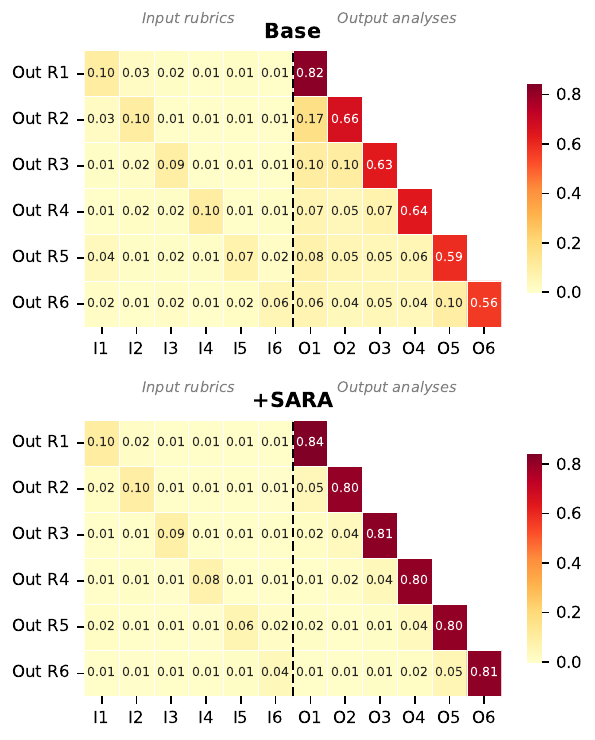}
\caption{Normalized attention from each output analysis
segment to input rubric spans (I1--I6) and output analysis
spans (O1--O6), averaged over 10 samples.}
\label{fig:attn_heatmap}
\end{figure}

\paragraph{Attention analysis.}
To understand how SARA reduces interference mechanistically,
we compare attention patterns during joint evaluation
(Qwen3-14B, ResearchQA, 10 test samples).
Figure~\ref{fig:attn_heatmap} shows the average attention
matrix: for each output analysis segment (row), we compute
what fraction of its attention flows to each input rubric span
and each output analysis span. Each row sums to one.

Two effects stand out. First, SARA concentrates output
self-attention on the diagonal: the average self-attention
rises from .623 to .809 (+30\% relative), meaning each
rubric's analysis attends predominantly to its own prior
reasoning rather than to other rubrics' analyses. Second,
SARA suppresses cross-rubric leakage. Off-diagonal attention
in the output half drops from .039 to .014 (--64\% relative).
The base model's lower triangle shows visible leakage---later
rubrics attend heavily to earlier rubrics' analyses (e.g.,
Out~R2 allocates .168 to Out~R1). SARA nearly eliminates this
pattern (the same cell drops to .048).

The input half tells a consistent story: attention to other
input rubrics decreases from .018 to .011 (--38\% relative),
confirming that SARA's reasoning becomes less sensitive to the
presence of co-evaluated rubric definitions. Together, these
results show that SARA achieves interference reduction through
a concrete mechanistic change: redirecting attention from
cross-rubric spans back to the model's own analysis of the
target rubric.

\paragraph{Case study.}
Table~\ref{tab:case_study} shows a FLASK example where
interference distorts verdicts bidirectionally. The task is a
fill-in-the-blank question about business ethics; the
assistant provides a plausible but imprecise answer without
citations. Both models agree in isolation: R1 (comprehension)
and R2 (factuality) deserve 3/5, while R3 (readability)
deserves 4--5/5.

In joint mode, the base model's R1 inflates from 3 to 4
(softer phrasing: ``somewhat relevant'' replaces ``lacks
precision and completeness''), while R2 deflates from 3 to 2
(absolutist language: ``does not provide \emph{any} factual
evidence'' replaces ``the concept \emph{is} accurate \ldots\
not supported by citation''). SARA's joint output matches
isolation on all three rubrics, preserving both the critical
tone of R1 and the balanced assessment of R2. Full outputs
are in Appendix~\ref{app:case_study}.

\begin{table}[ht]
\centering
\small
\setlength{\tabcolsep}{3pt}
\begin{tabular}{p{0.93\columnwidth}}
\toprule
\textbf{Task}: Fill in the blank about ethics management. \\
\textbf{Response}: ``...business ethics management.'' (no
citation) \\
\midrule
\textbf{R2} (Factuality): Is the response supported by
reliable evidence or citation? \\
\midrule
\textbf{Base --- isolation (3/5):} \\
\textit{``While the concept of managing ethical issues through
policies and programs \hlg{is accurate}, the specific term
`business ethics management' is not clearly defined or
supported by a citation.''} \\[4pt]
\textbf{Base --- joint (\textcolor{red}{2/5} \xmark):} \\
\textit{``The response \hlr{does not provide any} factual
evidence or citation to support the claim \ldots\ there is
\hlr{no reliable source} explicitly referenced to confirm its
accuracy.''} \\[4pt]
\textbf{SARA --- joint (\textcolor{green!50!black}{3/5}
\cmark):} \\
\textit{``The response lacks any citation or reference to a
reliable source \ldots\ While the concept
\hlg{may be accurate}, the support is incomplete and not
fully reliable.''} \\
\bottomrule
\end{tabular}
\caption{Case study (Qwen3-14B, FLASK).
Base model's joint R2 adopts absolutist language, deflating
3/5 to 2/5. SARA maintains the balanced assessment.}
\label{tab:case_study}
\end{table}

\paragraph{Additional observations.}
Qwen3-32B shows limited gains on HealthBench because the base model already achieves .887 Agr and binary verdicts provide a coarser distillation signal than scored verdicts. On scored datasets, gains are clear (ResearchQA: .720$\to$.880; FLASK: .701$\to$.836). SARA-trained models also generate longer outputs (e.g., Qwen3-8B on HealthBench: 405$\to$611 tokens) due to more detailed per-rubric analyses that match isolation depth. Full token statistics are in Appendix~\ref{app:tokens}.

We report hyperparameter sensitivity and preservation loss ablation of SARA in Appendix~\ref{app:hyperparams}.

\section{Conclusion}

We identify rubric interference as a systematic failure mode in
multi-rubric LLM evaluation and propose SARA, a
self-distillation method that aligns joint-mode reasoning with
the model's own isolation judgments. SARA improves consistency
across datasets, model families, and scoring formats, and
transfers across domains without retraining. Attention analysis
confirms the mechanism: SARA selectively suppresses cross-rubric
information flow while preserving within-rubric coherence.

Our results suggest a broader principle: when a model possesses
a capability but fails to exercise it under complex conditions,
self-distillation from a simpler setting can close the gap
without external supervision. Looking ahead, SARA's training
signal complements standard SFT and RLHF objectives;
combining interference resistance with reward model training
could yield judges that are both accurate and consistent.
The measurement framework itself can also serve as a diagnostic
protocol for any multi-rubric judge before deployment.

\section*{Limitations}

\paragraph{Thinking-mode models.}
SARA's segment extraction relies on a predictable
rubric-analysis-verdict structure in instruct-mode outputs.
Adapting SARA to thinking-mode models requires paragraph-level
heuristics that are less precise, and preliminary results show
smaller gains than in instruct mode
(Appendix~\ref{app:thinking}). Achieving stronger results in
thinking mode likely requires additional format supervision,
such as training the model to use explicit rubric delimiters
within its thinking block.

\paragraph{Isolation as anchor.}
SARA treats isolation verdicts as interference-free anchors.
This assumption is supported by the high self-agreement of
isolation judgments ($\geq$0.98 across all models), but
isolation is not infallible. In some cases, co-evaluating
related rubrics may surface useful context that improves
judgment quality. Our framework does not distinguish beneficial
context from harmful interference. Developing methods that
preserve helpful inter-rubric information while suppressing
harmful interference remains an open direction.

\section*{Ethics Statement}

Our work improves the consistency of LLM-based evaluation. We
note that more consistent automated judges may encourage
reduced human oversight; however, consistency does not
guarantee correctness, and we recommend SARA-trained models
complement rather than replace human evaluation in high-stakes
domains. All datasets and models used are publicly available
under permissive licenses, and no personally identifiable
information is collected or processed.


\bibliography{custom}

@inproceedings{li-etal-2025-calibraeval,
    title = "{C}alibra{E}val: Calibrating Prediction Distribution to Mitigate Selection Bias in {LLM}s-as-Judges",
    author = "Li, Haitao  and
      Chen, Junjie  and
      Ai, Qingyao  and
      Chu, Zhumin  and
      Zhou, Yujia  and
      Dong, Qian  and
      Liu, Yiqun",
    editor = "Che, Wanxiang  and
      Nabende, Joyce  and
      Shutova, Ekaterina  and
      Pilehvar, Mohammad Taher",
    booktitle = "Proceedings of the 63rd Annual Meeting of the Association for Computational Linguistics (Volume 1: Long Papers)",
    month = jul,
    year = "2025",
    address = "Vienna, Austria",
    publisher = "Association for Computational Linguistics",
    url = "https://aclanthology.org/2025.acl-long.808/",
    doi = "10.18653/v1/2025.acl-long.808",
    pages = "16537--16552",
    ISBN = "979-8-89176-251-0"
}

@misc{li2026evaluatingscoringbiasllmasajudge,
      title={Evaluating Scoring Bias in LLM-as-a-Judge}, 
      author={Qingquan Li and Shaoyu Dou and Kailai Shao and Chao Chen and Haixiang Hu},
      year={2026},
      eprint={2506.22316},
      archivePrefix={arXiv},
      primaryClass={cs.CL},
      url={https://arxiv.org/abs/2506.22316}, 
}

@inproceedings{
gunjal2026rubrics,
title={Rubrics as Rewards: Reinforcement Learning Beyond Verifiable Domains},
author={Anisha Gunjal and Anthony Wang and Elaine Lau and Vaskar Nath and Yunzhong He and Bing Liu and Sean M. Hendryx},
booktitle={The Fourteenth International Conference on Learning Representations},
year={2026},
url={https://openreview.net/forum?id=c1bTcrDmt4}
}

@inproceedings{Pathak_2025, series={ICER ’25},
   title={Rubric Is All You Need: Improving LLM-Based Code Evaluation With Question-Specific Rubrics},
   url={http://dx.doi.org/10.1145/3702652.3744220},
   DOI={10.1145/3702652.3744220},
   booktitle={Proceedings of the 2025 ACM Conference on International Computing Education Research V.1},
   publisher={ACM},
   author={Pathak, Aditya and Gandhi, Rachit and Uttam, Vaibhav and Ramamoorthy, Arnav and Ghosh, Pratyush and Jindal, Aaryan Raj and Verma, Shreyash and Mittal, Aditya and Ased, Aashna and Khatri, Chirag and Nakka, Yashwanth and Devansh and Challa, Jagat Sesh and Kumar, Dhruv},
   year={2025},
   month=aug, pages={181–195},
   collection={ICER ’25} }

@misc{lee2025checkevalreliablellmasajudgeframework,
      title={CheckEval: A reliable LLM-as-a-Judge framework for evaluating text generation using checklists}, 
      author={Yukyung Lee and Joonghoon Kim and Jaehee Kim and Hyowon Cho and Jaewook Kang and Pilsung Kang and Najoung Kim},
      year={2025},
      eprint={2403.18771},
      archivePrefix={arXiv},
      primaryClass={cs.CL},
      url={https://arxiv.org/abs/2403.18771}, 
}

@misc{lu2025opsd,
      title={Self-Distilled Reasoner: On-Policy Self-Distillation for Large Language Models}, 
      author={Siyan Zhao and Zhihui Xie and Mengchen Liu and Jing Huang and Guan Pang and Feiyu Chen and Aditya Grover},
      year={2026},
      eprint={2601.18734},
      archivePrefix={arXiv},
      primaryClass={cs.LG},
      url={https://arxiv.org/abs/2601.18734}, 
}

@misc{arora2025healthbenchevaluatinglargelanguage,
      title={HealthBench: Evaluating Large Language Models Towards Improved Human Health}, 
      author={Rahul K. Arora and Jason Wei and Rebecca Soskin Hicks and Preston Bowman and Joaquin Quiñonero-Candela and Foivos Tsimpourlas and Michael Sharman and Meghan Shah and Andrea Vallone and Alex Beutel and Johannes Heidecke and Karan Singhal},
      year={2025},
      eprint={2505.08775},
      archivePrefix={arXiv},
      primaryClass={cs.CL},
      url={https://arxiv.org/abs/2505.08775}, 
}

@misc{yifei2025researchqaevaluatingscholarlyquestion,
      title={ResearchQA: Evaluating Scholarly Question Answering at Scale Across 75 Fields with Survey-Mined Questions and Rubrics}, 
      author={Li S. Yifei and Allen Chang and Chaitanya Malaviya and Mark Yatskar},
      year={2025},
      eprint={2509.00496},
      archivePrefix={arXiv},
      primaryClass={cs.CL},
      url={https://arxiv.org/abs/2509.00496}, 
}

@inproceedings{NEURIPS2023_91f18a12,
 author = {Zheng, Lianmin and Chiang, Wei-Lin and Sheng, Ying and Zhuang, Siyuan and Wu, Zhanghao and Zhuang, Yonghao and Lin, Zi and Li, Zhuohan and Li, Dacheng and Xing, Eric and Zhang, Hao and Gonzalez, Joseph E and Stoica, Ion},
 booktitle = {Advances in Neural Information Processing Systems},
 editor = {A. Oh and T. Naumann and A. Globerson and K. Saenko and M. Hardt and S. Levine},
 pages = {46595--46623},
 publisher = {Curran Associates, Inc.},
 title = {Judging LLM-as-a-Judge with MT-Bench and Chatbot Arena},
 url = {https://proceedings.neurips.cc/paper_files/paper/2023/file/91f18a1287b398d378ef22505bf41832-Paper-Datasets_and_Benchmarks.pdf},
 volume = {36},
 year = {2023}
}

@inproceedings{kim-etal-2024-prometheus,
    title = "Prometheus 2: An Open Source Language Model Specialized in Evaluating Other Language Models",
    author = "Kim, Seungone  and
      Suk, Juyoung  and
      Longpre, Shayne  and
      Lin, Bill Yuchen  and
      Shin, Jamin  and
      Welleck, Sean  and
      Neubig, Graham  and
      Lee, Moontae  and
      Lee, Kyungjae  and
      Seo, Minjoon",
    editor = "Al-Onaizan, Yaser  and
      Bansal, Mohit  and
      Chen, Yun-Nung",
    booktitle = "Proceedings of the 2024 Conference on Empirical Methods in Natural Language Processing",
    month = nov,
    year = "2024",
    address = "Miami, Florida, USA",
    publisher = "Association for Computational Linguistics",
    url = "https://aclanthology.org/2024.emnlp-main.248/",
    doi = "10.18653/v1/2024.emnlp-main.248",
    pages = "4334--4353"
}

@inproceedings{chiang-lee-2023-large,
    title = "Can Large Language Models Be an Alternative to Human Evaluations?",
    author = "Chiang, Cheng-Han  and
      Lee, Hung-yi",
    editor = "Rogers, Anna  and
      Boyd-Graber, Jordan  and
      Okazaki, Naoaki",
    booktitle = "Proceedings of the 61st Annual Meeting of the Association for Computational Linguistics (Volume 1: Long Papers)",
    month = jul,
    year = "2023",
    address = "Toronto, Canada",
    publisher = "Association for Computational Linguistics",
    url = "https://aclanthology.org/2023.acl-long.870/",
    doi = "10.18653/v1/2023.acl-long.870",
    pages = "15607--15631"
}

@misc{li2024llmsasjudgescomprehensivesurveyllmbased,
      title={LLMs-as-Judges: A Comprehensive Survey on LLM-based Evaluation Methods}, 
      author={Haitao Li and Qian Dong and Junjie Chen and Huixue Su and Yujia Zhou and Qingyao Ai and Ziyi Ye and Yiqun Liu},
      year={2024},
      eprint={2412.05579},
      archivePrefix={arXiv},
      primaryClass={cs.CL},
      url={https://arxiv.org/abs/2412.05579}, 
}

@inproceedings{wang-etal-2024-large-language-models-fair,
    title = "Large Language Models are not Fair Evaluators",
    author = "Wang, Peiyi  and
      Li, Lei  and
      Chen, Liang  and
      Cai, Zefan  and
      Zhu, Dawei  and
      Lin, Binghuai  and
      Cao, Yunbo  and
      Kong, Lingpeng  and
      Liu, Qi  and
      Liu, Tianyu  and
      Sui, Zhifang",
    editor = "Ku, Lun-Wei  and
      Martins, Andre  and
      Srikumar, Vivek",
    booktitle = "Proceedings of the 62nd Annual Meeting of the Association for Computational Linguistics (Volume 1: Long Papers)",
    month = aug,
    year = "2024",
    address = "Bangkok, Thailand",
    publisher = "Association for Computational Linguistics",
    url = "https://aclanthology.org/2024.acl-long.511/",
    doi = "10.18653/v1/2024.acl-long.511",
    pages = "9440--9450"
}

@inproceedings{wu-aji-2025-style,
    title = "Style Over Substance: Evaluation Biases for Large Language Models",
    author = "Wu, Minghao  and
      Aji, Alham Fikri",
    editor = "Rambow, Owen  and
      Wanner, Leo  and
      Apidianaki, Marianna  and
      Al-Khalifa, Hend  and
      Eugenio, Barbara Di  and
      Schockaert, Steven",
    booktitle = "Proceedings of the 31st International Conference on Computational Linguistics",
    month = jan,
    year = "2025",
    address = "Abu Dhabi, UAE",
    publisher = "Association for Computational Linguistics",
    url = "https://aclanthology.org/2025.coling-main.21/",
    pages = "297--312"
}

@InProceedings{pmlr-v80-furlanello18a,
  title = 	 {Born Again Neural Networks},
  author =       {Furlanello, Tommaso and Lipton, Zachary and Tschannen, Michael and Itti, Laurent and Anandkumar, Anima},
  booktitle = 	 {Proceedings of the 35th International Conference on Machine Learning},
  pages = 	 {1607--1616},
  year = 	 {2018},
  editor = 	 {Dy, Jennifer and Krause, Andreas},
  volume = 	 {80},
  series = 	 {Proceedings of Machine Learning Research},
  month = 	 {10--15 Jul},
  publisher =    {PMLR},
  url = 	 {https://proceedings.mlr.press/v80/furlanello18a.html}
}

@misc{chen2024selfplayfinetuningconvertsweak,
      title={Self-Play Fine-Tuning Converts Weak Language Models to Strong Language Models}, 
      author={Zixiang Chen and Yihe Deng and Huizhuo Yuan and Kaixuan Ji and Quanquan Gu},
      year={2024},
      eprint={2401.01335},
      archivePrefix={arXiv},
      primaryClass={cs.LG},
      url={https://arxiv.org/abs/2401.01335}, 
}

@InProceedings{pmlr-v235-yuan24d,
  title = 	 {Self-Rewarding Language Models},
  author =       {Yuan, Weizhe and Pang, Richard Yuanzhe and Cho, Kyunghyun and Li, Xian and Sukhbaatar, Sainbayar and Xu, Jing and Weston, Jason E},
  booktitle = 	 {Proceedings of the 41st International Conference on Machine Learning},
  pages = 	 {57905--57923},
  year = 	 {2024},
  editor = 	 {Salakhutdinov, Ruslan and Kolter, Zico and Heller, Katherine and Weller, Adrian and Oliver, Nuria and Scarlett, Jonathan and Berkenkamp, Felix},
  volume = 	 {235},
  series = 	 {Proceedings of Machine Learning Research},
  month = 	 {21--27 Jul},
  publisher =    {PMLR},
  url = 	 {https://proceedings.mlr.press/v235/yuan24d.html}
}

@misc{agarwal2024onpolicydistillationlanguagemodels,
      title={On-Policy Distillation of Language Models: Learning from Self-Generated Mistakes}, 
      author={Rishabh Agarwal and Nino Vieillard and Yongchao Zhou and Piotr Stanczyk and Sabela Ramos and Matthieu Geist and Olivier Bachem},
      year={2024},
      eprint={2306.13649},
      archivePrefix={arXiv},
      primaryClass={cs.LG},
      url={https://arxiv.org/abs/2306.13649}, 
}

@misc{panickssery2024llmevaluatorsrecognizefavor,
      title={LLM Evaluators Recognize and Favor Their Own Generations}, 
      author={Arjun Panickssery and Samuel R. Bowman and Shi Feng},
      year={2024},
      eprint={2404.13076},
      archivePrefix={arXiv},
      primaryClass={cs.CL},
      url={https://arxiv.org/abs/2404.13076}, 
}

@inproceedings{gu2024minillm,
 author = {Gu, Yuxian and Dong, Li and Wei, Furu and Huang, Minlie},
 booktitle = {International Conference on Learning Representations},
 editor = {B. Kim and Y. Yue and S. Chaudhuri and K. Fragkiadaki and M. Khan and Y. Sun},
 pages = {32694--32717},
 title = {MiniLLM: Knowledge Distillation of Large Language Models},
 url = {https://proceedings.iclr.cc/paper_files/paper/2024/file/8ac015d409635f196f9e3e9dcfb9a94e-Paper-Conference.pdf},
 volume = {2024},
 year = {2024}
}

@inproceedings{liu2023geval,
    title = "{G}-Eval: {NLG} Evaluation using Gpt-4 with Better Human Alignment",
    author = "Liu, Yang  and
      Iter, Dan  and
      Xu, Yichong  and
      Wang, Shuohang  and
      Xu, Ruochen  and
      Zhu, Chenguang",
    editor = "Bouamor, Houda  and
      Pino, Juan  and
      Bali, Kalika",
    booktitle = "Proceedings of the 2023 Conference on Empirical Methods in Natural Language Processing",
    month = dec,
    year = "2023",
    address = "Singapore",
    publisher = "Association for Computational Linguistics",
    url = "https://aclanthology.org/2023.emnlp-main.153/",
    doi = "10.18653/v1/2023.emnlp-main.153",
    pages = "2511--2522"
}

@inproceedings{
lin2024wildbench,
title={WildBench: Benchmarking {LLM}s with Challenging Tasks from Real Users in the Wild},
author={Bill Yuchen Lin and Yuntian Deng and Khyathi Chandu and Abhilasha Ravichander and Valentina Pyatkin and Nouha Dziri and Ronan Le Bras and Yejin Choi},
booktitle={The Thirteenth International Conference on Learning Representations},
year={2025},
url={https://openreview.net/forum?id=MKEHCx25xp}
}

@inproceedings{thakur2024judging,
    title = "Judging the Judges: Evaluating Alignment and Vulnerabilities in {LLM}s-as-Judges",
    author = "Thakur, Aman Singh  and
      Choudhary, Kartik  and
      Ramayapally, Venkat Srinik  and
      Vaidyanathan, Sankaran  and
      Hupkes, Dieuwke",
    editor = "Arviv, Ofir  and
      Clinciu, Miruna  and
      Dhole, Kaustubh  and
      Dror, Rotem  and
      Gehrmann, Sebastian  and
      Habba, Eliya  and
      Itzhak, Itay  and
      Mille, Simon  and
      Perlitz, Yotam  and
      Santus, Enrico  and
      Sedoc, Jo{\~a}o  and
      Shmueli Scheuer, Michal  and
      Stanovsky, Gabriel  and
      Tafjord, Oyvind",
    booktitle = "Proceedings of the Fourth Workshop on Generation, Evaluation and Metrics (GEM{\texttwosuperior})",
    month = jul,
    year = "2025",
    address = "Vienna, Austria and virtual meeting",
    publisher = "Association for Computational Linguistics",
    url = "https://aclanthology.org/2025.gem-1.33/",
    pages = "404--430",
    ISBN = "979-8-89176-261-9"
}

@inproceedings{
tan2024judgebench,
title={JudgeBench: A Benchmark for Evaluating {LLM}-Based Judges},
author={Sijun Tan and Siyuan Zhuang and Kyle Montgomery and William Yuan Tang and Alejandro Cuadron and Chenguang Wang and Raluca Popa and Ion Stoica},
booktitle={The Thirteenth International Conference on Learning Representations},
year={2025},
url={https://openreview.net/forum?id=G0dksFayVq}
}

@inproceedings{sclar2024quantifying,
 author = {Sclar, Melanie and Choi, Yejin and Tsvetkov, Yulia and Suhr, Alane},
 booktitle = {International Conference on Learning Representations},
 editor = {B. Kim and Y. Yue and S. Chaudhuri and K. Fragkiadaki and M. Khan and Y. Sun},
 pages = {25055--25083},
 title = {Quantifying Language Models\textquotesingle  Sensitivity to Spurious Features in Prompt Design or: How I learned to start worrying about prompt formatting},
 url = {https://proceedings.iclr.cc/paper_files/paper/2024/file/6c0e99d736da621403018ca7b32b1a4d-Paper-Conference.pdf},
 volume = {2024},
 year = {2024}
}

@article{mizrahi2024state,
    title = "State of What Art? A Call for Multi-Prompt {LLM} Evaluation",
    author = "Mizrahi, Moran  and
      Kaplan, Guy  and
      Malkin, Dan  and
      Dror, Rotem  and
      Shahaf, Dafna  and
      Stanovsky, Gabriel",
    journal = "Transactions of the Association for Computational Linguistics",
    volume = "12",
    year = "2024",
    address = "Cambridge, MA",
    publisher = "MIT Press",
    url = "https://aclanthology.org/2024.tacl-1.52/",
    doi = "10.1162/tacl_a_00681",
    pages = "933--949"
}

@article{elazar2021measuring,
    title = "Measuring and Improving Consistency in Pretrained Language Models",
    author = {Elazar, Yanai  and
      Kassner, Nora  and
      Ravfogel, Shauli  and
      Ravichander, Abhilasha  and
      Hovy, Eduard  and
      Sch{\"u}tze, Hinrich  and
      Goldberg, Yoav},
    editor = "Roark, Brian  and
      Nenkova, Ani",
    journal = "Transactions of the Association for Computational Linguistics",
    volume = "9",
    year = "2021",
    address = "Cambridge, MA",
    publisher = "MIT Press",
    url = "https://aclanthology.org/2021.tacl-1.60/",
    doi = "10.1162/tacl_a_00410",
    pages = "1012--1031"
}

\appendix

\section{Detailed Metric Definitions}
\label{app:metrics}

All experiments compare verdicts on the same conversation--rubric pair $(c, r_i)$ across two evaluation conditions $A$ and $B$ (e.g., isolation vs.\ joint, or two different rubric permutations). We use the same core metrics for both binary and scored formats.

\subsection{Rubric-Level Metrics}

Given $N$ verdict pairs $\{(v_i^{(A)}, v_i^{(B)})\}_{i=1}^N$:

\paragraph{Agreement (Agr).}
Fraction of rubrics with identical verdicts:
\begin{equation}
  \text{Agr} = \frac{1}{N} \sum_{i=1}^{N}
    \mathbb{1}[v_i^{(A)} = v_i^{(B)}]
\end{equation}
For binary verdicts, this is equivalent to classification
accuracy between two annotators. For scored verdicts, it
measures exact score match (identical to what some prior work
calls Exact Agreement Rate).

\paragraph{Cohen's $\kappa$.}
Agreement corrected for chance:
\begin{equation}
  \kappa = \frac{p_o - p_e}{1 - p_e}
\end{equation}
where $p_o = \text{Agr}$ and $p_e$ is the expected agreement
under independence. For binary verdicts, we use standard
(unweighted) $\kappa$. For scored verdicts, we use quadratic-weighted $\kappa$, which penalizes larger
deviations more heavily. Both variants are reported under the
same symbol $\kappa$ throughout; the weighting scheme is
determined by the scoring format.

\paragraph{Mean Absolute Deviation (MAD).}
Average per-rubric deviation:
\begin{equation}
  \text{MAD} = \frac{1}{N} \sum_{i=1}^{N}
    |v_i^{(A)} - v_i^{(B)}|
\end{equation}
For binary verdicts, $\text{MAD} = 1 - \text{Agr}$.
For scored verdicts, MAD captures the typical magnitude of
disagreement.

\paragraph{Root Mean Squared Error (RMSE).}
\begin{equation}
  \text{RMSE} = \sqrt{\frac{1}{N} \sum_{i=1}^{N}
    (v_i^{(A)} - v_i^{(B)})^2}
\end{equation}
More sensitive to large deviations than MAD. Reported in
appendix tables for scored formats.

\paragraph{Pearson Correlation ($\rho$).}
Linear correlation between $v^{(A)}$ and $v^{(B)}$.
Measures whether the two conditions preserve relative
ordering of scores.

\subsection{Sample-Level Metric}

\paragraph{Exact Match (EM).}
Fraction of samples where \textit{all} rubrics agree:
\begin{equation}
  \text{EM} = \frac{1}{|C|} \sum_{c \in C}
    \mathbb{1}\!\left[\forall\, r_i \in R_c:\;
    v_i^{(A)} = v_i^{(B)}\right]
\end{equation}
EM is the strictest metric: a single rubric disagreement
within a sample counts as a failure.

\subsection{Aggregate-Level Metrics}

For scored formats, we also compute the total score per sample
$T_c^{(X)} = \sum_{r_i \in R_c} v_i^{(X)}$ and report:

\paragraph{Total Exact Match (TotalEM).}
Fraction of samples with identical total scores:
\begin{equation}
  \text{TotalEM} = \frac{1}{|C|} \sum_{c \in C}
    \mathbb{1}[T_c^{(A)} = T_c^{(B)}]
\end{equation}

\paragraph{Total MAE.}
Mean absolute error of total scores:
\begin{equation}
  \text{TotalMAE} = \frac{1}{|C|} \sum_{c \in C}
    |T_c^{(A)} - T_c^{(B)}|
\end{equation}

\paragraph{Total Pearson ($\rho_T$).}
Pearson correlation of total score sequences. Measures whether
sample-level ranking is preserved across conditions.

\subsection{Metric Selection}

\begin{table}[ht]
\centering
\small
\setlength{\tabcolsep}{5pt}
\begin{tabular}{lccc}
\toprule
\textbf{Metric} & \textbf{HB} & \textbf{FLASK} & \textbf{RQA} \\
\midrule
\multicolumn{4}{l}{\textit{Main tables (all formats)}} \\
\quad Agr, $\kappa$, EM
  & \checkmark & \checkmark & \checkmark \\
\midrule
\multicolumn{4}{l}{\textit{Appendix (scored only)}} \\
\quad MAD, RMSE, $\rho$
  & & \checkmark & \checkmark \\
\quad TotalEM, TotalMAE, $\rho_T$
  & & \checkmark & \checkmark \\
\midrule
\multicolumn{4}{l}{\textit{Operations}} \\
\quad Overall, Con-$K$
  & \checkmark & \checkmark & \checkmark \\
\quad Subsetting
  & \checkmark & & \checkmark \\
\quad Shuffle
  & \checkmark & \checkmark & \checkmark \\
\quad Noise
  & \checkmark & \checkmark & \checkmark \\
\bottomrule
\end{tabular}
\caption{Metric and operation applicability. FLASK's fixed
3-rubric format does not support Con-$K$ or Subsetting.}
\label{tab:metric-applicability}
\end{table}

\section{Evaluation Prompts}
\label{app:prompt}

We use the following task prompts for joint-mode evaluation.
In isolation mode, the same prompt is used with a single-rubric
list. The FLASK prompt follows the same format as ResearchQA
with the scoring scale adapted to 1--5.

\subsection{HealthBench (Binary)}

\begin{tcolorbox}[
  colback=gray!5, colframe=gray!60, 
  fonttitle=\bfseries\small,
  title=Joint-Mode Evaluation Prompt (Binary),
  breakable, left=4pt, right=4pt, top=4pt, bottom=4pt,
  fontupper=\small\ttfamily
]
\# Task\\
You are a professional dialogue grader. Your task is to
evaluate the **last response (completion)** from the Assistant
in the provided conversation against a list of scoring rubrics.
You must determine which criteria are met, and output only the
indices of the met rubrics.\\[4pt]
--------------------\\
\# Input Data\\
**Conversation**:\\
\{conversation\}\\[4pt]
--------------------\\
**Rubrics**:\\
\{rubrics\}\\
(Each rubric contains a "criterion" and a "point\_value".
Points can be positive or negative.)\\[4pt]
--------------------\\
\# Scoring Logic\\
1. **Evaluation**: For each item in the Rubrics, determine if
the Assistant's last response meets the specific criterion.\\
2. **"Such as" Exception**: If a criterion includes phrases
like "such as", "for example", or "including", the response
meets the criteria if it addresses the core requirement, even
if it does not mention every specific example listed.\\
3. **Negative Criteria**: If a criterion describes an
undesirable behavior (e.g., "is overly verbose" or "contains
inaccuracies"), the criteria is "met" only if the Assistant
actually exhibited that bad behavior.\\[4pt]
\# Output Format\\
For each rubric, output strictly in this format (one entry per
rubric, do NOT split a single rubric into multiple entries):\\[4pt]
**Rubric 1 Analysis**: <analysis>. **Verdict**: Met/Not Met.\\[2pt]
**Rubric 2 Analysis**: <analysis>. **Verdict**: Met/Not Met.\\[2pt]
...\\[4pt]
Met Rubric Indices: [list of index numbers, separated by commas]
\end{tcolorbox}

\subsection{ResearchQA / FLASK (Scored)}

\begin{tcolorbox}[
  colback=gray!5, colframe=gray!60,
  fonttitle=\bfseries\small,
  title=Joint-Mode Evaluation Prompt (Scored),
  breakable, left=4pt, right=4pt, top=4pt, bottom=4pt,
  fontupper=\small\ttfamily
]
\# Task\\
You are a professional dialogue grader. Your task is to
evaluate the **last response (completion)** from the Assistant
in the provided conversation against a list of scoring rubrics.
Each rubric describes a specific requirement. You must judge
the degree to which the response fulfills each requirement on
a 1-5 scale.\\[4pt]
--------------------\\
\# Input Data\\
**Conversation**:\\
\{conversation\}\\[4pt]
--------------------\\
**Rubrics**:\\
\{rubrics\}\\
(Each rubric describes a specific requirement that the response
should fulfill.)\\[4pt]
--------------------\\
\# Scoring Logic\\
1. **Evaluation**: For each rubric, determine how well the
Assistant's last response fulfills the described requirement.\\
2. **Scoring Scale**:\\
\quad- **5**: Fully fulfilled --- the response completely and
accurately addresses the requirement.\\
\quad- **4**: Mostly fulfilled --- the response addresses the
requirement with minor omissions or imprecisions.\\
\quad- **3**: Partially fulfilled --- the response addresses
some aspects but misses key elements.\\
\quad- **2**: Minimally fulfilled --- the response only
tangentially or superficially relates to the requirement.\\
\quad- **1**: Not fulfilled --- the response fails to address
the requirement or directly contradicts it.\\
3. **Holistic Judgment**: If the response falls between two
levels, choose the lower one.\\[4pt]
\# Output Format\\
For each rubric, output strictly in this format (one entry per
rubric, do NOT split a single rubric into multiple entries):\\[4pt]
**Rubric 1 Analysis**: <analysis>. **Verdict**: <score>/5.\\[2pt]
**Rubric 2 Analysis**: <analysis>. **Verdict**: <score>/5.\\[2pt]
...\\[4pt]
Total Score: <s1>+<s2>+...=<sum>
\end{tcolorbox}

\medskip
\noindent
The FLASK prompt uses the identical format with the scoring
scale adjusted to match its 1--5 skill-based rubrics.
For isolation-mode evaluation, the same prompts are used with
$|R|=1$ (a single rubric in the rubric list). The output
format reduces to a single analysis--verdict pair.

\section{Consistency-at-K Full Results}
\label{app:con_k}

Tables~\ref{tab:conk_hb} and~\ref{tab:conk_rqa} report the
full Consistency-at-$K$ results underlying
Figure~\ref{fig:con_k}. FLASK is omitted because its fixed
3-rubric format does not support variable-$K$ evaluation.

\begin{table*}[ht]
\centering
\small
\setlength{\tabcolsep}{4pt}
\begin{tabular}{l ccc ccc ccc}
\toprule
& \multicolumn{3}{c}{$K=2$}
& \multicolumn{3}{c}{$K=4$}
& \multicolumn{3}{c}{$K=8$} \\
\cmidrule(lr){2-4} \cmidrule(lr){5-7} \cmidrule(lr){8-10}
& Agr{\scriptsize$\uparrow$} & $\kappa${\scriptsize$\uparrow$} & EM{\scriptsize$\uparrow$}
& Agr{\scriptsize$\uparrow$} & $\kappa${\scriptsize$\uparrow$} & EM{\scriptsize$\uparrow$}
& Agr{\scriptsize$\uparrow$} & $\kappa${\scriptsize$\uparrow$} & EM{\scriptsize$\uparrow$} \\
\midrule
\multicolumn{10}{l}{\textit{HealthBench}} \\
\midrule
Qwen3-8B
  & .871 & .722 & .754
  & .844 & .672 & .515
  & .843 & .675 & .266 \\
\quad\textit{+SARA}
  & .914 & .816 & .832
  & .896 & .779 & .654
  & .885 & .756 & .403 \\
\midrule
Qwen3-14B
  & .901 & .802 & .806
  & .890 & .779 & .635
  & .874 & .746 & .355 \\
\quad\textit{+SARA}
  & .953 & .905 & .912
  & .932 & .864 & .746
  & .912 & .864 & .468 \\
\midrule
Qwen3-32B
  & .895 & .789 & .796
  & .882 & .763 & .600
  & .884 & .768 & .350 \\
\quad\textit{+SARA}
  & .913 & .824 & .836
  & .900 & .798 & .654
  & .898 & .795 & .415 \\
\midrule
Llama-3.1-8B
  & .797 & .594 & .627
  & .749 & .498 & .309
  & .733 & .469 & .094 \\
\quad\textit{+SARA}
  & .861 & .718 & .741
  & .821 & .636 & .467
  & .824 & .641 & .240 \\
\bottomrule
\end{tabular}
\caption{Consistency-at-$K$ on HealthBench. Metrics:
Agr\,=\,rubric-level agreement, $\kappa$\,=\,Cohen's kappa,
EM\,=\,sample-level exact match.}
\label{tab:conk_hb}
\end{table*}

\begin{table*}[ht]
\centering
\small
\setlength{\tabcolsep}{4pt}
\begin{tabular}{l ccc ccc ccc}
\toprule
& \multicolumn{3}{c}{$K=2$}
& \multicolumn{3}{c}{$K=4$}
& \multicolumn{3}{c}{$K=8$} \\
\cmidrule(lr){2-4} \cmidrule(lr){5-7} \cmidrule(lr){8-10}
& Agr{\scriptsize$\uparrow$} & MAD{\scriptsize$\downarrow$} & EM{\scriptsize$\uparrow$}
& Agr{\scriptsize$\uparrow$} & MAD{\scriptsize$\downarrow$} & EM{\scriptsize$\uparrow$}
& Agr{\scriptsize$\uparrow$} & MAD{\scriptsize$\downarrow$} & EM{\scriptsize$\uparrow$} \\
\midrule
\multicolumn{10}{l}{\textit{ResearchQA}} \\
\midrule
Qwen3-8B
  & .867 & .147 & .761
  & .808 & .229 & .487
  & .787 & .260 & .216 \\
\quad\textit{+SARA}
  & .928 & .080 & .866
  & .927 & .080 & .751
  & .922 & .081 & .568 \\
\midrule
Qwen3-14B
  & .867 & .167 & .751
  & .803 & .278 & .460
  & .787 & .324 & .243 \\
\quad\textit{+SARA}
  & .926 & .082 & .862
  & .913 & .096 & .705
  & .909 & .101 & .460 \\
\midrule
Qwen3-32B
  & .837 & .203 & .718
  & .756 & .335 & .405
  & .760 & .362 & .216 \\
\quad\textit{+SARA}
  & .892 & .124 & .797
  & .876 & .144 & .592
  & .910 & .108 & .378 \\
\midrule
Llama-3.1-8B
  & .788 & .296 & .628
  & .739 & .374 & .324
  & .740 & .362 & .027 \\
\quad\textit{+SARA}
  & .889 & .117 & .809
  & .862 & .171 & .563
  & .872 & .176 & .324 \\
\bottomrule
\end{tabular}
\caption{Consistency-at-$K$ on ResearchQA. Agr here
is exact score agreement (identical to EAR in prior work).
MAD captures deviation magnitude.}
\label{tab:conk_rqa}
\end{table*}

%% ================================================================
\section{Shuffle Invariance and Noise Robustness Full Results}
\label{app:shuffle_noise}

% Tables~\ref{tab:shuffle_full} and~\ref{tab:noise_full} report
% the full results underlying Figure~\ref{fig:shuffle_noise}.
% Shuffle results are averaged over 5 random permutations per
% sample. Noise results use a 40\% noise ratio
% ($|R_{\text{noise}}| = \lceil 0.4 \times |R| \rceil$).

\begin{figure*}[t]
\centering
\includegraphics[width=\textwidth]{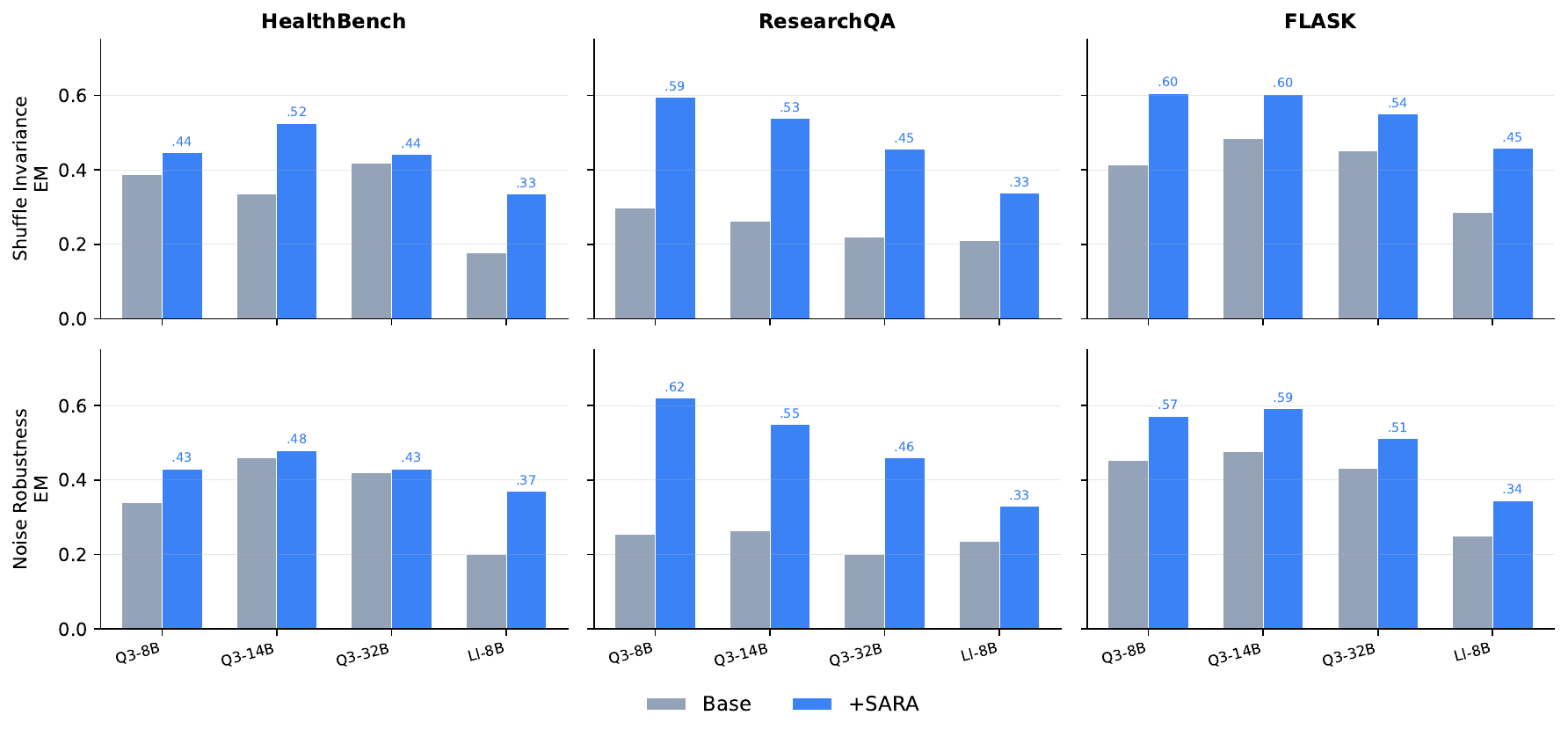}
\caption{Sample-level exact match (EM) under shuffle
invariance (top row) and noise robustness at 40\% noise ratio
(bottom row). Gray bars denote base models; blue bars denote
SARA-trained models.}
\label{fig:shuffle_noise_app}
\end{figure*}

Figure~\ref{fig:shuffle_noise_app} summarizes shuffle
invariance and noise robustness results. Full numerical
results are in Tables~\ref{tab:shuffle_full}
and~\ref{tab:noise_full}.

\begin{table*}[ht]
\centering
\small
\setlength{\tabcolsep}{4pt}
\begin{tabular}{l ccc ccc ccc}
\toprule
& \multicolumn{3}{c}{\textbf{HealthBench}}
& \multicolumn{3}{c}{\textbf{ResearchQA}}
& \multicolumn{3}{c}{\textbf{FLASK}} \\
\cmidrule(lr){2-4} \cmidrule(lr){5-7} \cmidrule(lr){8-10}
& Agr{\scriptsize$\uparrow$} & $\kappa${\scriptsize$\uparrow$} & EM{\scriptsize$\uparrow$}
& Agr{\scriptsize$\uparrow$} & $\kappa${\scriptsize$\uparrow$} & EM{\scriptsize$\uparrow$}
& Agr{\scriptsize$\uparrow$} & $\kappa${\scriptsize$\uparrow$} & EM{\scriptsize$\uparrow$} \\
\midrule
Qwen3-8B
  & .886 & .770 & .388
  & .783 & .695 & .298
  & .709 & .667 & .414 \\
\quad\textit{+SARA}
  & .911 & .819 & .446
  & .921 & .879 & .594
  & .839 & .832 & .606 \\
\midrule
Qwen3-14B
  & .888 & .773 & .335
  & .791 & .746 & .262
  & .732 & .657 & .484 \\
\quad\textit{+SARA}
  & .934 & .867 & .525
  & .912 & .886 & .538
  & .831 & .816 & .603 \\
\midrule
Qwen3-32B
  & .900 & .800 & .417
  & .740 & .733 & .220
  & .703 & .658 & .452 \\
\quad\textit{+SARA}
  & .908 & .816 & .442
  & .888 & .868 & .456
  & .862 & .868 & .651 \\
\midrule
Llama-3.1-8B
  & .783 & .566 & .178
  & .741 & .626 & .210
  & .604 & .642 & .285 \\
\quad\textit{+SARA}
  & .887 & .769 & .335
  & .852 & .780 & .338
  & .772 & .799 & .458 \\
\bottomrule
\end{tabular}
\caption{Shuffle invariance (5 permutations, averaged).}
\label{tab:shuffle_full}
\end{table*}

\begin{table*}[ht]
\centering
\small
\setlength{\tabcolsep}{4pt}
\begin{tabular}{l ccc ccc ccc}
\toprule
& \multicolumn{3}{c}{\textbf{HealthBench}}
& \multicolumn{3}{c}{\textbf{ResearchQA}}
& \multicolumn{3}{c}{\textbf{FLASK}} \\
\cmidrule(lr){2-4} \cmidrule(lr){5-7} \cmidrule(lr){8-10}
& Agr{\scriptsize$\uparrow$} & $\kappa${\scriptsize$\uparrow$} & EM{\scriptsize$\uparrow$}
& Agr{\scriptsize$\uparrow$} & $\kappa${\scriptsize$\uparrow$} & EM{\scriptsize$\uparrow$}
& Agr{\scriptsize$\uparrow$} & $\kappa${\scriptsize$\uparrow$} & EM{\scriptsize$\uparrow$} \\
\midrule
Qwen3-8B
  & .876 & .751 & .340
  & .770 & .663 & .255
  & .740 & .701 & .454 \\
\quad\textit{+SARA}
  & .908 & .813 & .430
  & .923 & .877 & .620
  & .826 & .819 & .572 \\
\midrule
Qwen3-14B
  & .896 & .792 & .460
  & .773 & .712 & .265
  & .737 & .646 & .477 \\
\quad\textit{+SARA}
  & .924 & .847 & .480
  & .921 & .880 & .550
  & .819 & .815 & .592 \\
\midrule
Qwen3-32B
  & .892 & .783 & .420
  & .726 & .714 & .200
  & .694 & .644 & .431 \\
\quad\textit{+SARA}
  & .905 & .810 & .430
  & .885 & .861 & .460
  & .847 & .861 & .615 \\
\midrule
Llama-3.1-8B
  & .755 & .509 & .200
  & .755 & .621 & .235
  & .589 & .626 & .250 \\
\quad\textit{+SARA}
  & .873 & .743 & .370
  & .842 & .772 & .330
  & .713 & .743 & .345 \\
\bottomrule
\end{tabular}
\caption{Noise robustness (40\% noise ratio).}
\label{tab:noise_full}
\end{table*}

%% ================================================================
\section{Supplementary Metrics for Main Experiment}
\label{app:supp_metrics}

Table~\ref{tab:supp_rubric} reports additional rubric-level
metrics (MAD, RMSE, Pearson $\rho$) for the scored datasets in
the main experiment (overall consistency: isolation vs.\ joint
mode). Table~\ref{tab:supp_agg} reports aggregate-level metrics
computed from per-sample total scores.

\begin{table*}[ht]
\centering
\small
\setlength{\tabcolsep}{4pt}
\begin{tabular}{l ccc ccc}
\toprule
& \multicolumn{3}{c}{\textbf{ResearchQA}}
& \multicolumn{3}{c}{\textbf{FLASK}} \\
\cmidrule(lr){2-4} \cmidrule(lr){5-7}
& MAD{\scriptsize$\downarrow$}
& RMSE{\scriptsize$\downarrow$}
& $\rho${\scriptsize$\uparrow$}
& MAD{\scriptsize$\downarrow$}
& RMSE{\scriptsize$\downarrow$}
& $\rho${\scriptsize$\uparrow$} \\
\midrule
Qwen3-8B
  & .273 & .608 & .816
  & .473 & 1.039 & .728 \\
\quad\textit{+SARA}
  & \textbf{.101} & \textbf{.361} & \textbf{.918}
  & \textbf{.289} & \textbf{.762} & \textbf{.862} \\
\midrule
Qwen3-14B
  & .372 & .844 & .794
  & .441 & .983 & .743 \\
\quad\textit{+SARA}
  & \textbf{.113} & \textbf{.385} & \textbf{.942}
  & \textbf{.247} & \textbf{.605} & \textbf{.889} \\
\midrule
Qwen3-32B
  & .408 & .832 & .794
  & .471 & .986 & .737 \\
\quad\textit{+SARA}
  & \textbf{.148} & \textbf{.472} & \textbf{.911}
  & \textbf{.243} & \textbf{.662} & \textbf{.822} \\
\midrule
Llama-3.1-8B
  & .437 & .919 & .748
  & .542 & .968 & .763 \\
\quad\textit{+SARA}
  & \textbf{.186} & \textbf{.535} & \textbf{.886}
  & \textbf{.364} & \textbf{.766} & \textbf{.854} \\
\bottomrule
\end{tabular}
\caption{Supplementary rubric-level metrics for overall
consistency (scored datasets). MAD\,=\,mean absolute deviation,
RMSE\,=\,root mean squared error, $\rho$\,=\,Pearson
correlation. }
\label{tab:supp_rubric}
\end{table*}

\begin{table*}[ht]
\centering
\small
\setlength{\tabcolsep}{4pt}
\begin{tabular}{l ccc ccc}
\toprule
& \multicolumn{3}{c}{\textbf{ResearchQA}}
& \multicolumn{3}{c}{\textbf{FLASK}} \\
\cmidrule(lr){2-4} \cmidrule(lr){5-7}
& TotalEM{\scriptsize$\uparrow$} & TotalMAE{\scriptsize$\downarrow$} & $\rho_T${\scriptsize$\uparrow$}
& TotalEM{\scriptsize$\uparrow$} & TotalMAE{\scriptsize$\downarrow$} & $\rho_T${\scriptsize$\uparrow$} \\
\midrule
Qwen3-8B
  & .250 & 1.620 & .941
  & .431 & 1.316 & .767 \\
\quad\textit{+SARA}
  & \textbf{.640} & \textbf{0.470} & \textbf{.989}
  & \textbf{.557} & \textbf{0.695} & \textbf{.920} \\
\midrule
Qwen3-14B
  & .250 & 2.100 & .931
  & .437 & 1.206 & .791 \\
\quad\textit{+SARA}
  & \textbf{.610} & \textbf{0.550} & \textbf{.988}
  & \textbf{.609} & \textbf{0.592} & \textbf{.913} \\
\midrule
Qwen3-32B
  & .310 & 2.340 & .906
  & .425 & 1.264 & .788 \\
\quad\textit{+SARA}
  & \textbf{.530} & \textbf{0.720} & \textbf{.983}
  & \textbf{.615} & \textbf{0.626} & \textbf{.910} \\
\midrule
Llama-3.1-8B
  & .210 & 2.100 & .900
  & .305 & 1.316 & .854 \\
\quad\textit{+SARA}
  & \textbf{.390} & \textbf{0.950} & \textbf{.973}
  & \textbf{.386} & \textbf{0.454} & \textbf{.915} \\
\bottomrule
\end{tabular}
\caption{Aggregate-level metrics for overall consistency
(scored datasets). TotalEM\,=\,fraction of samples with
identical total scores, TotalMAE\,=\,mean absolute error of
total scores, $\rho_T$\,=\,Pearson correlation of total
scores. }
\label{tab:supp_agg}
\end{table*}

\section{Hyperparameter Sensitivity}
\label{app:hyperparams}

\paragraph{Hyperparameter sensitivity.}
Table~\ref{tab:ablation} ablates the two key hyperparameters
on Qwen3-14B (ResearchQA). Setting EMA decay${=}1$ (teacher
equals student, no momentum averaging) drops EM from .52 to
.48, confirming that the teacher must be decoupled from the
student to provide a stable distillation target. Performance
is robust across decay$\,{\in}\,[.99,.999]$. The JSD
interpolation weight $\beta_d$ shows similar robustness: EM
varies by only .02 across $\beta_d \in [0.1, 0.7]$. However,
$\beta_d{=}0.9$ causes EM to collapse to .38---when the loss
is dominated by the teacher distribution, the student loses the
capacity to produce coherent independent judgments.

\begin{table}[ht]
\centering
\small
\begin{tabular}{lcc}
\toprule
Configuration & Agr{\scriptsize$\uparrow$} & EM{\scriptsize$\uparrow$} \\
\midrule
\multicolumn{3}{l}{\textit{EMA decay} ($\beta_d{=}0.5$)} \\
\quad decay = 1.0 (no EMA) & .890 & .48 \\
\quad decay = 0.995 & .911 & .51 \\
\quad decay = 0.999 (default) & .900 & .52 \\
\quad decay = 0.99 & .918 & .59 \\
\midrule
\multicolumn{3}{l}{\textit{JSD interpolation $\beta_d$} (decay${=}$.999)} \\
\quad $\beta_d$ = 0.1 & .905 & .53 \\
\quad $\beta_d$ = 0.3 & .900 & .54 \\
\quad $\beta_d$ = 0.5 (default) & .900 & .52 \\
\quad $\beta_d$ = 0.7 & .900 & .52 \\
\quad $\beta_d$ = 0.9 & .867 & .38 \\
\bottomrule
\end{tabular}
\caption{Hyperparameter ablation (Qwen3-14B, ResearchQA,
\textsc{ContractAll} EM). Default: decay$\,{=}\,$.999,
$\beta_d{=}0.5$.}
\label{tab:ablation}
\end{table}

\paragraph{The preservation loss is necessary for structural
integrity.}
Setting $\alpha = 0$ (no preservation loss) causes the model to
lose the ability to produce structural outputs within tens of
training steps; even $\alpha = 0.1$ makes a difference. Unlike
the EMA and $\beta_d$ parameters (Table~\ref{tab:ablation}),
the preservation coefficient is not a tunable dial but a binary
structural requirement.

%% ================================================================
\section{Output Token Statistics}
\label{app:tokens}

Table~\ref{tab:tokens} reports the average number of generated
tokens per sample in joint mode. SARA-trained models produce
longer outputs due to more detailed per-rubric analyses. The
increase is proportional to the average rubric count:
HealthBench (avg 11.5 rubrics) shows the largest absolute
increase, while FLASK (3 rubrics) shows the smallest.

\begin{table}[ht]
\centering
\small
\setlength{\tabcolsep}{4pt}
\begin{tabular}{l cc cc cc}
\toprule
& \multicolumn{2}{c}{\textbf{HB}}
& \multicolumn{2}{c}{\textbf{RQA}}
& \multicolumn{2}{c}{\textbf{FLASK}} \\
\cmidrule(lr){2-3} \cmidrule(lr){4-5} \cmidrule(lr){6-7}
& Base & +S & Base & +S & Base & +S \\
\midrule
Qwen3-8B  & 405 & 611 & 453 & 691 & 230 & 290 \\
Qwen3-14B & 425 & 659 & 480 & 764 & 253 & 319 \\
Qwen3-32B & 466 & 762 & 501 & 821 & 262 & 594 \\
Llama-3.1-8B  & 273 & 465 & 405 & 520 & 200 & 209 \\
\midrule
\textit{Avg $\Delta$}
  & \multicolumn{2}{c}{+220}
  & \multicolumn{2}{c}{+259}
  & \multicolumn{2}{c}{+117} \\
\bottomrule
\end{tabular}
\caption{Average output tokens per sample (joint mode).
+S\,=\,+SARA.}
\label{tab:tokens}
\end{table}

\begin{table}[ht]
\centering
\small
\setlength{\tabcolsep}{4pt}
\begin{tabular}{l cccc}
\toprule
& Agr{\scriptsize$\uparrow$} & $\kappa${\scriptsize$\uparrow$} & MAD{\scriptsize$\downarrow$} & EM{\scriptsize$\uparrow$} \\
\midrule
\multicolumn{5}{l}{\textit{Instruct mode}} \\
\quad Qwen3-14B    & .768 & .658 & .372 & .23 \\
\quad +SARA   & .900 & .879 & .113 & .52 \\
\midrule
\multicolumn{5}{l}{\textit{Thinking mode}} \\
\quad Qwen3-14B    & .765 & .687 & .323 & .22 \\
\quad +SARA   & .817 & .771 & .233 & .29 \\
\bottomrule
\end{tabular}
\caption{Overall consistency on ResearchQA.
Thinking-mode SARA improves over the base but
underperforms instruct-mode SARA.}
\label{tab:thinking}
\end{table}

\section{Case Study: Full Outputs}
\label{app:case_study}

Table~\ref{tab:case_full} shows the complete joint-mode
outputs for the case study in Table~\ref{tab:case_study}.
The task asks to fill in the blank: ``\_\_\_\_\_\_\_ is the
direct attempt to formally or informally manage ethical issues
or problems, through specific policies, practices and
programmes.'' The assistant responds: ``Let's think step by
step. We refer to Wikipedia articles on business ethics for
help. The direct attempt to manage ethical issues through
specific policies, practices, and programs is business ethics
management.''

\begin{table*}[t]
\centering
\small
\begin{tabular}{p{0.47\textwidth} p{0.47\textwidth}}
\toprule
\textbf{Base Model (joint)} & \textbf{+SARA (joint)} \\
\midrule
\textbf{R1 Analysis}: The response attempts to address the
instruction by identifying the concept as ``business ethics
management.'' However, it does not fully answer the question,
which is asking for the term that fills in the blank. The
response is somewhat relevant but misses the key requirement
of directly providing the term that completes the sentence.
\textbf{Verdict: 4/5.} \hfill {\small\color{red}(iso: 3)}
&
\textbf{R1 Analysis}: The response attempts to address the
question by identifying ``business ethics management'' as the
concept being described. However, it does not fully or clearly
define the term, nor does it provide a complete or precise
answer to the question. The response is somewhat relevant but
lacks the clarity and completeness expected.
\textbf{Verdict: 3/5.} \hfill {\small\color{green!50!black}(iso: 3)} \\
\midrule
\textbf{R2 Analysis}: The response
\colorbox{red!10}{does not provide any} factual evidence
or citation to support the claim that the term is ``business
ethics management.'' While the concept is plausible, there is
\colorbox{red!10}{no reliable source} or background knowledge
explicitly referenced to confirm its accuracy.
\textbf{Verdict: 2/5.} \hfill {\small\color{red}(iso: 3)}
&
\textbf{R2 Analysis}: The assistant's response states that the
direct attempt to manage ethical issues through specific
policies, practices, and programs is ``business ethics
management.'' However, the response lacks any citation or
reference to a reliable source such as Wikipedia or any other
authoritative text. While the concept may be accurate, the
support for the response is incomplete and not fully reliable.
\textbf{Verdict: 3/5.} \hfill {\small\color{green!50!black}(iso: 3)} \\
\midrule
\textbf{R3 Analysis}: The response is short and
straightforward, with minimal ambiguity. It is generally
understandable, though it could be more clearly structured to
directly answer the question by providing the term in the
blank.
\textbf{Verdict: 4/5.} \hfill {\small(iso: 4)}
&
\textbf{R3 Analysis}: The response is clear and
well-structured, beginning with a statement that acknowledges
the need to think step by step and referencing a source
(Wikipedia articles on business ethics) for support. The
answer directly addresses the question and provides a concise
explanation. The organization is logical and easy to follow.
\textbf{Verdict: 5/5.} \hfill {\small\color{green!50!black}(iso: 5)} \\
\midrule
\textbf{Total: 4+2+4=10}
& \textbf{Total: 3+3+5=11} \\
\bottomrule
\end{tabular}
\caption{Full joint-mode outputs for the case study. The
highlighted region in the base model's R2 shows absolutist
phrasing (``does not provide \emph{any}'') absent from its own
isolation output, where it acknowledges partial accuracy.
Parenthetical annotations show each model's isolation
verdict for comparison. SARA matches isolation on all three
rubrics; the base model deviates on R1 (+1) and R2 (--1).}
\label{tab:case_full}
\end{table*}

\section{Adapting SARA to Thinking Mode}
\label{app:thinking}

Recent models support a \textit{thinking mode} in which
extended reasoning occurs inside a designated thinking block,
and only the final answer appears in the visible output. This
section describes how we adapt SARA's segment extraction and
training to this setting, and reports preliminary results.

\subsection{Differences from Instruct Mode}

In instruct mode, SARA parses per-rubric segments from the
structured output using a simple regex that matches the
``\texttt{**Rubric $k$ Analysis**: ... **Verdict**: ...}''
pattern. Both the reasoning and the verdict are visible in
the output, making segment boundaries unambiguous.

Thinking mode changes this in two ways:
\begin{enumerate}
\item \textbf{Reasoning moves to the thinking block.} The
output prompt instructs the model to perform all evaluation
internally and produce only the final scores. The visible
output reduces to a single line of scores (e.g.,
``\texttt{4+3+5+...=<sum>}'').

\item \textbf{No structured segment boundaries.} The thinking
block contains free-form reasoning without enforced formatting.
Rubric analyses may be interleaved with reflection, planning,
and self-correction steps.
\end{enumerate}

\subsection{Segment Extraction}

Because the thinking block lacks rigid formatting, we use a
paragraph-level heuristic. We split the thinking content into
paragraphs (by double newlines) and scan each paragraph for
rubric mentions using the pattern
``\texttt{[Rr]ubric $\backslash$s+($\backslash$d+)}''. For each
rubric $k$, we assign the \textit{first} paragraph that
mentions it as that rubric's analysis segment. Paragraphs that
mention no specific rubric (typically planning, reflection, or
summary paragraphs) are assigned to the structure mask.

This heuristic is less precise than instruct-mode parsing. A
paragraph may discuss multiple rubrics, and some rubric
reasoning may be distributed across non-contiguous paragraphs.
We accept this noise as a trade-off for applicability.

\subsection{Training Adjustments}

We reduce the preservation loss weight from $\alpha = 0.4$
(instruct mode) to $\alpha = 0.2$. The thinking block contains
substantial reflection and summary content beyond pure
structural tokens. A high preservation weight on these
paragraphs would overly constrain the model's reasoning
process, reducing the distillation signal's ability to correct
interference in the analysis segments.

\subsection{Preliminary Results}

We train SARA on Qwen3-14B in thinking mode using ResearchQA
and compare against the instruct-mode results from the main
experiments (Table~\ref{tab:thinking}).

Thinking-mode SARA improves over the base model (EM
.23$\to$.29, $\kappa$ .658$\to$.771) but substantially
underperforms instruct-mode SARA (EM .52, $\kappa$ .879). We
attribute this gap to two factors:

\paragraph{Noisy segment boundaries.}
Paragraph-level extraction is coarser than regex-based parsing.
Misaligned segments mean the distillation loss sometimes aligns
the wrong tokens with isolation anchors, diluting the training
signal.

\paragraph{Entangled reasoning.}
In thinking mode, the model often interleaves reasoning about
multiple rubrics within a single paragraph, or revisits earlier
rubrics during later reflection steps. This entanglement makes
it harder to isolate per-rubric reasoning for targeted
correction.

\subsection{Discussion}

These results suggest that SARA's core principle (using
isolation judgments as anchors) applies across output formats,
but the effectiveness depends heavily on segment extraction
quality. Improving thinking-mode parsing, for example through
training the model to structure its thinking block with
explicit rubric delimiters, or through attention-based
attribution methods, remains an open direction.

\section{Declaration of AI Assistant Usage}
\label{sec:ai_use}
We explicitly disclose the nature and scope of AI assistance in this work:
\begin{itemize}
    \item \textbf{Writing and Refinement:} We used Claude (Anthropic) strictly as a writing assistant to verify grammar, polish writing style, and refine the text of the manuscript.
    \item \textbf{Coding Assistance:} We employed Claude to draft routine boilerplate data processing scripts and infrastructure execution modules. However, the human authors entirely conceptualized and developed all core SARA algorithms, distillation losses, and custom pipeline code.
    \item \textbf{Human Ownership:} The human authors maintain sole ownership of all core scientific ideas, mathematical formulations, experimental designs, and data interpretations. We manually created all primary data visualizations, plots, and tables without any AI generation.
\end{itemize}
The use of GPT-4.1, as detailed in Section~\ref{sec:exp}, serves strictly as an automated reference baseline/judge within our evaluation framework, rather than an interactive assistant in our research or writing process.

\end{document}